\RequirePackage{fix-cm}
\documentclass[11pt]{article}

\usepackage{fullpage}
\usepackage{graphicx}

\usepackage{multirow}
\usepackage{amsmath,amsfonts,amssymb}
\usepackage{algpseudocode}
\usepackage[utf8]{inputenc}
\usepackage{array}
\usepackage{graphicx,color,enumerate}
\usepackage{hyperref}
\usepackage{graphicx}
\usepackage{varwidth}
\usepackage[ruled]{algorithm}
\usepackage{epstopdf}

\usepackage{capt-of}

\newcommand{\OO}{\mathcal{O}}

\newcommand{\JJ}{\mathcal{J}}
\newcommand{\MM}{\mathcal{M}}

\newcommand{\RR}{\mathbb{R}}

\newcommand{\CC}{{\bf C}}

\newcommand{\coarsen}{\textsf{coarsen}$-\bullet$}
\newcommand{\uncoarsen}{\textsf{uncoarsen}$-\bullet$}
\newcommand{\refine}{\textsf{refine}$-\bullet$}

\newcommand{\mlsvm}{\textsf{mlsvm}}
\newcommand{\sv}{\textsf{sv}}

\usepackage{graphicx}
\usepackage{calc}
\newlength{\depthofsumsign}
\newlength{\totalheightofsumsign}
\newlength{\heightanddepthofargument}

\makeatletter
\newcommand*{\DivideLengths}[2]{%
  \strip@pt\dimexpr\number\numexpr\number\dimexpr#1\relax*65536/\number\dimexpr#2\relax\relax sp\relax
}
\makeatother

\def\NoNumber#1{{\def\alglinenumber##1{}\State #1}\addtocounter{ALG@line}{-1}}

\newcommand{\mb}[1]{\mbox{\boldmath $#1$}}

\newcommand{\bx}{\mb{x}}

\usepackage{changes}
\definechangesauthor[color=blue]{is}
\definechangesauthor[color=red]{es}
\definechangesauthor[color=orange]{tl}

\begin{document}

\title{Engineering fast multilevel support vector machines}

\author{Ehsan Sadrfaridpour$^1$, Talayeh Razzaghi$^2$, Ilya Safro$^1$\footnote{Corresponding author isafro@clemson.edu}
\vspace{.3cm}\\
%
1- School of Computing, Clemson University, Clemson SC, USA
%
\vspace{.1cm}\\
2- Department of Industrial Engineering, New Mexico State University, Las Cruces NM, USA 
\vspace{.1cm}\\
}

\maketitle


%
%

\date{}

\maketitle
\begin{abstract}
The computational complexity of solving nonlinear support vector machine (SVM) is prohibitive  on large-scale data. In particular, this issue becomes very sensitive when the data represents additional difficulties such as highly imbalanced class sizes. Typically, nonlinear kernels produce significantly higher classification quality to linear kernels but introduce extra kernel and model parameters which requires computationally expensive fitting. This increases the quality but also reduces the performance dramatically. We introduce a generalized fast multilevel framework for regular and weighted SVM and discuss several versions of its algorithmic components that lead to a good trade-off between quality and time. Our framework is implemented using PETSc which allows an easy integration with scientific computing tasks. 
The experimental results demonstrate significant speed up compared to the state-of-the-art nonlinear SVM libraries.\\
Reproducibility: our source code, documentation and parameters are available at \url{https://github.com/esadr/mlsvm}.\\

\end{abstract}
\section{Introduction} \label{intro} 
Support vector machine (SVM) is one of the most well-known supervised classification methods that has been extensively used in such fields as disease diagnosis, text categorization, and fraud detection. Training nonlinear SVM classifier (such as Gaussian kernel based) requires solving convex quadratic programming (QP) model whose running time can be prohibitive for large-scale instances without using specialized acceleration techniques such as sampling, boosting, and hierarchical training. Another typical reason of increased running time is complex data sets (e.g., when the data is noisy, imbalanced, or incomplete) that require using model selection techniques for finding the best model parameters.

The motivation behind this work was extensive applied experience with hard, large-scale, industrial (often highly heterogeneous) data sets for which fast \emph{linear} SVMs produced extremely low quality results (as well as many other fast methods), and various nonlinear SVMs exhibited a strong trade off between running time and quality. It has been noticed in multiple works that many different real-world data sets have a strong underlying multiscale (in some works called hierarchical) structure \cite{KushnirGB06,karypis1999chameleon,lee2009convolutional,sharon06Hierarchy} that can be discovered through careful definitions of coarse-grained resolutions. Not surprisingly, we found that among fast methods the hierarchical nonlinear SVM was the best candidate for producing most satisfying results in a reasonable time \cite{asharaf2006scalable}. 
Although, several successful hierarchical SVM techniques \cite{yu2003classifying,hao2007hierarchically} have been developed since massive popularization of SVM, we found that most existing algorithms do not sustainably produce high-quality results in a short running time, and the behavior of hierarchical training is still not well studied. This is in contrast to a variety of well studied unsupervised multiscale clustering approaches \cite{brandes2008modularity,noack2009multi,rotta2011multilevel}. 

In this paper, we discuss several techniques for engineering multilevel SVMs demonstrating their (dis)advantages and generalizing them in a framework inspired by the algebraic multigrid and multiscale optimization strategies \cite{vlsicad}. We deliberately omit the issues related to parallelization of multilevel frameworks as it has been discussed in a variety of works related to multilevel clustering, partitioning, and SVM QP solvers. Our goal is to demonstrate fast and scalable sequential techniques focusing on different aspects of building and using multilevel learning with regular and weighted SVM. Also, we focus only on nonlinear SVMs because (a) not much improvement can be introduced or required in practice to accelerate linear SVMs, and (b) in many hard practical cases, the quality of linear SVMs is incomparable to that of nonlinear SVMs. The most promising and stable version of our multilevel SVMs are implemented in PETSc \cite{petsc-user-ref} which is a well known scientific computing library. PETSc was selected because of its scalability of linear algebra computations on large sparse matrices and available software infrastructure for future parallelization. Our implementation also addresses a critical need \cite{ascr-report-ml15} of adding data analysis functionality to broadly used scientific computing software.

\subsection{Computational challenges}
There is a number of basic challenges one has to address when applying SVM which we successfully tackle with the multilevel framework, namely, QP solver complexity for large-scale data, imbalanced data, and SVM model parameter fitting.
%
%
%
\paragraph{Large-scale data}
%
The baseline SVM classifier is typically formulated as a convex QP problem whose solvers scale between $\OO(n^2)$ to $\OO(n^3)$ \cite{graf2004parallel}. 
For example, the solver we compare our algorithm with, namely, LibSVM \cite{chang2011libsvm}, 
which is one of the most popular and fast QP solvers, scales between $\OO(n_{f}{n_{s}}^2)$ to $\OO(n_{f}{n_{s}}^3)$ subject to how effectively the cache is exploited in practice, where $n_f$ and $n_s$ are the number of features and samples, respectively. Clearly, this complexity is prohibitive for nonlinear SVM models applied on practical big data without using parallelization, high-performance computing systems or another special treatment.

\paragraph{Imbalanced data} 
The imbalanced data is one of the issues in which SVM often outperforms many fast machine learning methods.  This problem occurs when the number of instances of one class (negative or majority class) is substantially larger than the number of instances that belong to the other class (positive or minority class). In multi-class classification, the problem of imbalanced data is even bolder and use of the standard classification methods become problematic in the presence of big and imbalanced data \cite{lopez2015cost}. This may dramatically deteriorate the performance of the algorithm. It is worth noticing that there are cases in which correct classification of the smaller class is more important than misclassification of the larger class \cite{sun2007cost}. 
Fault diagnosis \cite{yang2009association,zhu2010fault}, anomaly detection \cite{khreich2010iterative,tavallaee2010toward}, medical diagnosis \cite{mazurowski2008training} are some of applications which are known to suffer of this problem.
Imbalanced data was one of our motivating factors because we have noticed that most standard SVM solvers do not behave well on it. 
In order to reduce the effect of minority class misclassification in highly imbalanced data, an extension of SVM, namely, the cost-sensitive SVM (whose extensions are also known as weighted or fuzzy SVM)  \cite{lin2002fuzzy}, was developed for imbalanced classification problems. In cost-sensitive SVM, a special control of misclassification penalization is introduced as a part of the SVM model.
%
%
%
%
\paragraph{Parameter tuning}
The quality of SVM models is very sensitive to the parameters (such as penalty factors of misclassified data) especially in case of using kernels that typically introduce extra parameters. There are many different parameter tuning approaches such as \cite{bao2013pso,zhou2009credit,lin2008parameter,chapelle2002choosing,cawley2010over,an2007fast,luts2010tutorial,lessmann2006genetic}. 
However, in any case, tuning parameters requires multiple executions of the training process for different parameters and due to the k-fold cross-validation which significantly increases the running time of the entire framework. In our experiments with industrial and healthcare data, not surprisingly, we were unable to find an acceptable quality SVM models without parameter fitting (also known as model selection  \cite{coussement2008churn,zhang2010aco,scholkopf2002learning}) which also motivated our work.
%
%
%
\subsection{Related work}
Multiple approaches have been proposed to improve the performance of SVM solvers.
Examples include efficient serial algorithms that use a cohort of decomposition techniques \cite{osuna1997improved}, shrinking and caching \cite{joachims1999making}, and fast second order working set selection \cite{fan2005working}. 
A popular LibSVM solver \cite{chang2011libsvm} implements the sequential minimal optimization  algorithm. 
In the cases of simple data for which nonlinear SVM is not required such approaches as LibLINEAR \cite{fan2008liblinear} demonstrate excellent performance for linear SVM using a coordinate descent algorithm which is very fast but, typically, not suitable for complex or imbalanced data.
Another approach to accelerate the QP solvers is a chunking \cite{joachims1999making}, in which the models are solved iteratively on the subsets of training data until the global optimum is achieved.

A typical acceleration of support vector machines is done through parallelization and training on high-performance computing systems using interior-point methods (IPM) \cite{mehrotra1992implementation} applied on the dual problem which is a convex QP. The key idea of the primal-dual IPM is to remove inequality constraints using a barrier function and then resort to the iterative Newton’s method to solve the KKT system of the dual problem. For example, in PSVM \cite{zhu2008parallelizing}, the algorithm reduces memory use, and parallelizes data loading and computation in IPM. It improves the decomposition-based LibSVM from $O(n^2)$ to $O(np^2/m)$, where $m$ is a number of processors (or heterogeneous machines used), and $p$ is a column dimension of a factorized matrix that is required for effective distribution of the data. The HPSVM solver  \cite{li2016hpsvm} is also based on solving the primal-dual IPM and uses effective parallelizm of factorization. The approach is specifically designed to take maximal advantage of the CPU-GPU collaborative computation with the dual buffers 3-stage pipeline mechanism, and efficiently handles large-scale training datasets. In HPSVM, the heterogeneous hierarchical memory is explored to optimize the bottleneck of data transfer. The P-packSVM \cite{zhu2009p} parallelizes the stochastic gradient descent solver of SVM that directly optimizes the
primal objective with the help of a distributed hash table and sophisticated data packing strategy. Other works utilize many-core GPUs to accelerate the sequential minimal optimization \cite{platt1999fast}, and other architectures  \cite{you2015scaling}.

One of the most well known works in which hierarchical SVM technique was introduced to improve the performance and quality of a classifier is \cite{yu2003classifying}. The coarsesning consists of creating a hierarchical clustered representation of the data points that are merged pairwise using Euclidean distance criterion. In this work, only linear classifiers are discussed and no inheritance and refinement of model parameters was introduced. A similar hierarchical clustering framework was proposed for non-linear SVM kernels in combination with feature selection techniques to develop an advanced intrusion detection system \cite{Horng2011306}.
Another coarsening approach that uses k-means clustering was introduced in 
\cite{pmlr-v32-hsieha14}. \emph{In all these works, the quality of classifiers strictly depends on how well the data is clustered using a particular clustering method applied on it.} Our coarsening scheme is more gradual and flexible than the clustering methods in these papers. Most of them, however, can be generalized as algebraic multigrid restriction operators (will be discussed further) in special forms. Also, in our frameworks, we emphasize several important aspects of training such as coarse level models, imbalanced coarsening, and parameter learning that are typically not considered in hierarchical SVM frameworks.

Multilevel Divide-and-Conquer SVM (DC-SVM) was developed using adaptive clustering and early prediction strategy \cite{pmlr-v32-hsieha14}. It outperforms previously mentioned methods, so we compare the computational performance and quality of classification for both DC-SVM and our proposed framework. 
The training time of DC-SVM for \emph{a fixed set of parameters} is fast. However, in order to achieve high quality classifiers a parameter fitting is typically required.
While DC-SVM with parameter fitting is faster than state-of-the-art \emph{exact} SVMs, it is significantly slower than our proposed framework.
Our experimental results (that include the parameter fitting component) show significant performance improvement on benchmark data sets in comparison to DC-SVM. 


In several works, a scalable parallelization of hierarchical SVM frameworks is developed to minimize the communication \cite{you2015svm,graf2004parallel,Cui2017}. Such techniques can be used on top of our framework. Successful results obtained using hierarchical structures have been shown specifically for multi-class classification \cite{cheong2004support,hao2007hierarchically,Khan:2007:NID:1285882.1285904,puget2015hierarchical}. 
Another relevant line of research is related to multilevel clustering and segmentation methods  \cite{kushnir2006fast,fang2010multilevel,sharon06Hierarchy}. They produce solutions at different levels of granularity which makes them suitable for visualization, aggregation of data, and building a hierarchical solution.

\subsection{Multilevel algorithmic frameworks}
In this paper, we discuss a practical construction of multilevel algorithmic frameworks (MAF) for SVM. These frameworks are inspired by the multiscale optimization strategies \cite{vlsicad}.
(We note that there exist several frameworks termed multilevel SVMs. These, however, correspond to completely different ideas. We preserve the terminology of multilevel, and multiscale optimization  algorithms.)
The main objective of multilevel algorithms is to construct a hierarchy of problems (coarsening), each approximating the original problem but with fewer degrees of freedom. This is achieved by introducing a chain of successive restrictions of the problem domain into low-dimensional or smaller-size domains and solving the coarse problems in them using local processing (uncoarsening) \cite{raey,DhillonGuanKulis05fast}. The MAF combines solutions obtained by the local processing at different levels of coarseness into one global solution. Such frameworks have several key advantages that make them  attractive for applying on large-scale data: they typically exhibit linear complexity (see Sec. \ref{sec:complexity}),
and are relatively easily parallelized. Another advantage
of the MAF is its heterogeneity, expressed in the ability to incorporate external appropriate optimization algorithms (as a refinement) in the framework at different levels. For example, if some SVM model selection technique is found to be particularly successful in parameter finding and obtaining high-quality solutions on some class of datasets, one can incorporate this technique at all levels of MAF and accelerate it by 1) applying it locally, 2) combining local solutions into global, and 3) inheriting parameters trained at coarse levels. 
These frameworks are extremely successful in various practical machine learning tasks such as clustering \cite{mlmodul}, segmentation \cite{sharon06Hierarchy}, and dimensionality reduction \cite{raey}.

The major difference between typical computational optimization MAF, and those that we introduce for SVM is the output of the model. 
In SVM, the main output is the set of the support vectors which is usually much smaller at all levels of the multilevel hierarchy than the total number of data points at the corresponding levels. We use this observation in our methods by redefining the training set during the uncoarsening and making MAF scalable. In particular, we inherit the support vectors from the coarse scales, add their neighborhoods, and refine the support vectors at all scales. 
In other words, we improve the separating hyperplane throughout the hierarchy by gradual refinement of the support vectors until a global solution at the finest level is reached.
In addition, we inherit the parameters of model selection and kernel from the coarse levels, and refine them throughout the uncoarsening.

\subsection{Our contribution}
We introduce novel methods of engineering fast and high quality multilevel frameworks for efficient and effective training of nonlinear SVM classifiers. We also summarize and generalize existing \cite{razzaghi2015scalable,razzaghi2016multilevel} approaches. We discuss various coarsening strategies, and introduce the weighted aggregation framework inspired by the algebraic multigrid \cite{vlsicad} which significantly improves and generalizes all of them. In the weighted aggregation framework, the data points are either partitioned in hierarchical fashion where small groups of data points are aggregated or split into fractions where different fractions of the same data point can belong to different aggregates.  
Without any notable loss in the quality of classifiers, multilevel SVM frameworks exhibit substantially faster running times and are able to generate \emph{several} classifiers at  different coarse-grained resolutions in one complete training iteration 
which also helps to interpret these classifiers qualitatively (see Section \ref{sec:mod-scales}). Depending on the size and structure of the training set, the resulting final decision rule of our multilevel classifier will be either exactly the same as in single SVM model or composed as voting of several smaller SVM models.

The proposed multilevel frameworks are particularly effective on imbalanced data sets where fitting model parameters is the most computationally expensive component. Our multilevel frameworks can be parallelized as any algebraic multigrid  algorithm and their superiority is demonstrated on several publicly available and industrial data sets.
The performance improvement over the best sequential state-of-the-art nonlinear SVM libraries with high classification quality is significant.
For example, on the average, for large data sets we boost the performance 491 times over LibSVM and 45 times over the DC-SVM (which was chosen because of its superiority over other hierarchical methods mentioned above). On some large datasets, a full comparison was impossible because of infeasible running time of the competitive approaches which demonstrates superiority of the proposed method.


\section{Preliminaries}
We define the optimization problems underlying SVM models for binary classification. 
Given a set $\JJ$ that contains $n$ data points $x_i \in \RR^d$, $1\leq i\leq n$, we define the corresponding labeled pairs $(x_i,y_i)$, where each $x_i$ belongs to the class determined by a given label $y_i \in \{-1, 1\}$.
 Data points with positive labels are called the \emph{minority} class which is denoted by $\CC^+$ with $|\CC^+| = n^+$. The rest of the points belongs to the \emph{majority} class which is 
denoted by $\CC^-$, where $|\CC^-| = n^-$, i.e., $\JJ=\CC^+ \cup \CC^-$.
Solving the following convex optimization problem by finding $w$, and $b$ produces a hyperplane
with maximum margin between $\CC^+$, and $\CC^-$
\begin{align}\label{eqn:SVM}
\textsf{minimize} & &  \frac{1}{2}\lVert w \rVert^2 + {C}\sum_{i=1}^n \xi_i \\
\textsf{subject to}  & & y_i (w^T \phi(x_i) + b) \geq 1 - \xi_i, &\quad i = 1, \dots,n \nonumber\\
& & \xi_i \geq 0, &\quad i=1, \dots, n \nonumber.
\end{align}
The mapping of data points to higher dimensional space is done by $\phi:\RR^d \rightarrow \RR^p ~ (d \leq p)$ to make two classes separable by a hyperplane.
The term slack variables $\{\xi_i\}_{i=1}^n$ are used to penalize misclassified points.
The parameter $C > 0$ controls the magnitude of the penalization. 
The primal formulation is shown at (\ref{eqn:SVM}) which is known as the \textit{soft margin} SVM  \cite{wu2005svm}.


The weighted SVM (WSVM) addresses imbalanced problems with assigning different weights to classes with parameters $C^+$ and $C^-$. The set of slack variables is split into two disjoint sets $\{\xi_i^+\}_{i=1}^{n^+}$, and $\{\xi_i^-\}_{i=1}^{n^-}$, respectively.
In WSVM, the objective of (\ref{eqn:SVM}) is changed into 
\begin{align}\label{eqn:WSVM}
  \textsf{minimize} ~~~~~ & \frac{1}{2}\lVert w \rVert^2 + {C^+}\sum_{i=1}^{n^+}  \xi_i^+ + {C^-}\sum_{j=1}^{n^-}  \xi_j^-. 
\end{align}

Solving the Lagrangian dual problem using kernel functions $k(\bx_i,\bx_j) = {\phi(x_i)}^T \phi(x_j)$ produces a reliable convergence which is faster than methods for primal formulations (\ref{eqn:SVM}) and (\ref{eqn:WSVM}). In our framework, we use the sequential minimal optimization solver implemented in LibSVM library~\cite{chang2011libsvm}. 
The role of kernel functions is to measure the similarity for pairs of points $x_i$ and $x_j$. We present computational results with the Gaussian kernel (RBF), $\exp(-\gamma ||x_i-x_j||^2)$, which is known to be generally reliable  
when no additional assumptions about the data are known. 
Experiments with other kernels exhibit improvements that are similar to those with RBF if compared with regular  (W)SVM solver with the same kernels. Technically, using another kernel requires only switching to it in the refinement at the uncoarsening stage (see Alg. \ref{alg:iisref}) including parameter inheritance, if required. We note that some of our experimental datasets are not solved well with non-RBF kernels used in regular (W)SVM solver, so here we demonstrate the results only for RBF.

In order to achieve an acceptable quality of the classifier, many difficult data sets require reinforcement of (W)SVM with tuning methods for such model parameters as $C$, $C^+$, $C^-$, and kernel function parameters (e.g., the bandwidth parameter $\gamma$ for RBF kernel function). This is one of the major sources of running time complexity of (W)SVM models which we are aiming to improve.

In our framework we use the adapted nested uniform design (NUD) model selection algorithm to fit the parameters \cite{huang2007model} which is a popular model selection technique for (W)SVM. The main intuition behind NUD is that it finds the close-to-optimal parameter set in an iterative nested manner. The optimal solution is calculated in terms of maximizing the required performance measure (such as accuracy and G-mean). 
Although, we study binary classification problems, it can easily be extended to the multi-class classification using either directed multi-class classification or transforming the problem into multiple independent binary (W)SVMs that can be processed independently in parallel. 

%
%
\paragraph{Two-level problem}
In order to describe the (un)coarsening algorithms, we introduce the two-level problem notation that can be extended into full multilevel hierarchy (see Figure \ref{fig:svm}). We will use subscript $(\cdot)_f$ and $(\cdot)_c$ to represent fine and coarse variables, respectively. For example, the data points of two consecutive levels, namely, fine and coarse, will be denoted by $\JJ_f$, and $\JJ_c$, respectively. The sets of fine and coarse support vectors are denoted by $\sv_f$, and $\sv_c$, respectively. We will also use a subscript in the parentheses to denote the level number in the hierarchy where appropriate. For example, $\JJ_{(i)}$ will denote the set of data points at level $i$.

\paragraph{Proximity graphs}
All multilevel (W)SVM frameworks discussed in subsequent sections are based on different coarsening schemes for creating a hierarchy of data proximity graphs. Initially, at the finest level, $\JJ$ is represented as two $k$-nearest neighbor ($k$NN) graphs $G^+_{(0)} = (\CC^+, E^+)$, and $G^-_{(0)} = (\CC^-, E^-)$ for minority and majority classes, respectively, where each $x_i\in \CC^{+(-)}$ corresponds to a node in $G^{+(-)}_{(0)}$. A pair of nodes in $G_{(0)}^{+(-)}$ is connected with an edge that belongs to $E^{+(-)}$ if one of them belongs to a set of $k$-nearest neighbors of another. In practice, we are using \emph{approximate} $k$-nearest neighbors graphs (A$k$NN) as our experiments with the \emph{exact} nearest neighbor graphs do not demonstrate any improvement in the quality of classifiers whereas computing them is a time consuming task. In the computational experiments, we used FLANN library  \cite{muja_flann_2009,muja2014scalable}. Results obtained  with other approximate nearest neighbor search algorithms are found to be not significantly different. 
Throughout the multilevel hierarchies, in two-level representation, the fine and coarse level graphs will be denoted by $G^{+(-)}_f = (\CC^{+(-)}_f, E^{+(-)}_f)$, and $G^{+(-)}_c = (\CC^{+(-)}_c, E^{+(-)}_c)$, respectively. All coarse graphs, except $G^{+(-)}_{(0)}$ are obtained using respective coarsening algorithm.

\paragraph{Multiple models} In the proposed multilevel frameworks, when the data is too big, independent training of several subsets of the data will be performed. As a result, a training on $k$ subsets will produce $k$ models that will be denoted  as $\{(\sv_f, C^+_f, C^-_f, \gamma_f)_i\}_{i=1}^k$ to avoid introducing additional index for each parameter.
\section{Multilevel support vector machines}

\begin{figure}
\centering
\includegraphics[width=.8\textwidth]{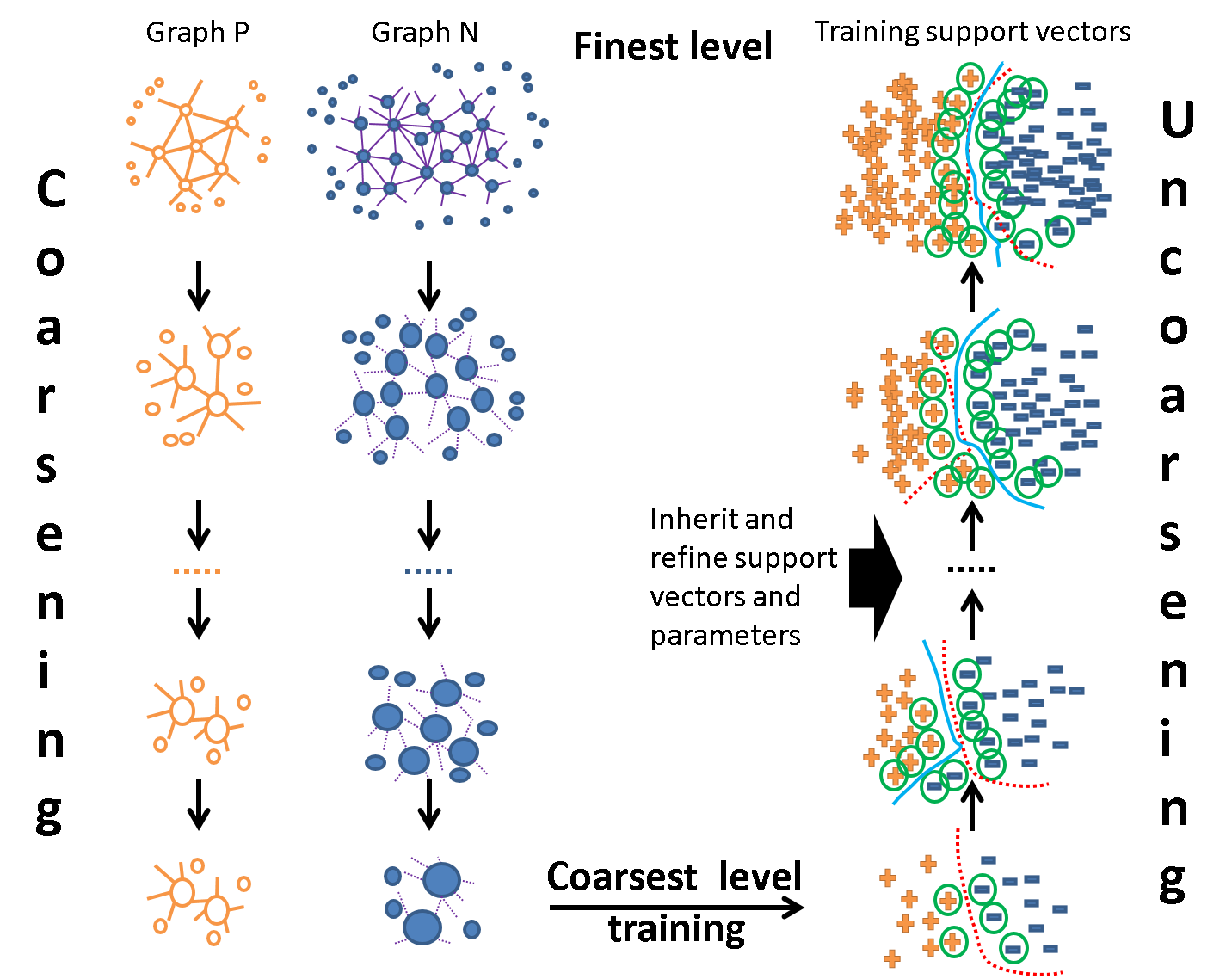}
\caption{Multilevel SVM coarsening-uncoarsening framework scheme.}\label{fig:svm}
\end{figure}

The multilevel frameworks discussed in this paper include three phases (see Figure \ref{fig:svm}), namely,  gradual training set coarsening, coarsest support vector learning, and gradual support vector refinement (uncoarsening). In the training set coarsening phase, we create a hierarchy of coarse training set representations, $\JJ_{(i)}$, in which each next-coarser level ($i+1$) contains a fewer number of points than in the previous level ($i$) such that the coarse level learning problem approximates the fine level problem. The coarse level training points are not necessarily the same fine level points (such as in \cite{razzaghi2015scalable}) or their strict small clusters (such as in \cite{yu2003classifying}). 

When the size of training set is sufficiently small to apply a high quality training algorithm for given computational resources, the set of coarsest support vectors and model parameters are trained. We denote by $M^{+(-)}$ the upper limit for the sizes of coarsest training sets which should depend on the ability of available computational resources to solve the problem exactly in a reasonable time. In the uncoarsening, both the support vectors and model parameters are inherited from the coarse level and improved using local refinement at the fine level. The uncoarsening is continued from the coarsest to the finest levels as is shown in Figure \ref{fig:svm}. Separate coarsening hierarchies are created for classes $\CC^+$, and $\CC^-$, independently. 

The main driving routine, \mlsvm-$\bullet$, of a multilevel (W)SVM framework is presented in Algorithm \ref{alg:dr}. The SVM cost-sensitive framework is designed similarly with a parameter $C$, see Eq. (\ref{eqn:SVM}). In Algorithm \ref{alg:dr}, the functions \coarsen, \uncoarsen, and \refine~ are the building blocks of the multilevel framework discussed in this paper. These functions will differ from multilevel framework to framework. The bullet ``$\bullet$'' will be replaced with corresponding method names.

%
%


\subsection{Iterative Independent Set Multilevel Framework}

We describe the coarsening only for class $\CC_f^+$ as the same process works for $\CC_f^-$. The multilevel framework (\mlsvm-IIS, Alg. \ref{alg:dr}) with iterative independent set coarsening applies several iterative passes in each of which a set of fine points is selected and added to the set of coarse points $\CC^+_c$. In order to cover the space of points uniformly, this is done by selecting independent sets of nodes in $G^+_f$. The independent set is a set of vertices in a graph whose node-induced subgraph has no edges. We present this coarsening in details in \cite{razzaghi2015scalable}.

\begin{algorithm}[H]
\caption{\mlsvm-$\bullet$($\CC^+_f, \CC^-_f, G^+_f, G^-_f, M^+, M^-$): multilevel (W)SVM main driving routine. The functions \coarsen, \uncoarsen, and \refine~ are the building blocks of the multilevel framework. They will differ from multilevel framework to framework. ``$\bullet$'' will be replaced by method names described in following sections.}
\label{alg:dr}
\begin{algorithmic}[1]
\State \textbf{if} {$|\JJ_f| \leq M^+ + M^-$} \textbf{then}  \Comment{Solve the problem exactly if the data is small}
\State \quad $(\sv_f, C^+, C^-, \gamma) \leftarrow $ train (W)SVM model on $\JJ_f$ (including NUD)\label{ln:cst}  
\State \textbf{else }  \Comment{Create and solve a coarse problem. Then refine its solution at level $f$.}
\State \quad \textbf{if} {$|\CC^+_f| \leq M^+$} \textbf{then} $\CC^+_c \leftarrow \CC^+_f;~ G^+_c \leftarrow G^+_f$
\State \quad \quad \textbf{else} $(\CC^+_c, G^+_c) \leftarrow$ \coarsen$(\CC^+_f, G^+_f)$
\State \quad \textbf{if} {$|\CC^-_f| \leq M^-$} \textbf{then} $\CC^-_c \leftarrow \CC^-_f;~ G^-_c \leftarrow G^-_f$
\State \quad \quad \textbf{else} $(\CC^-_c, G^-_c) \leftarrow$ \coarsen$(\CC^-_f, G^-_f)$
\State \quad $(\sv_c, \tilde{C}^+, \tilde{C}^-, \tilde{\gamma})\leftarrow$ \mlsvm-$\bullet$($\CC^+_c, \CC^-_c$, $G^+_c$, $G^-_c$, $M^+$, $M^-$)
\State \quad $\sv_f\leftarrow$ \uncoarsen$(\sv_c)$ \Comment{Project support vectors from $c$ to $f$ and add neighbors}
\NoNumber{\Comment{Get one or more ($k$) models if the data is too big at current level}}
\State \quad $\{(\sv_f, C^+, C^-, \gamma)_i\}_{i=1}^k \leftarrow$  \refine$( \sv_f, \tilde{C}^+, \tilde{C}^-, \tilde{\gamma})$ 
\State  \textbf{if} $f$ is the finest level \textbf{then}
\State \quad \textbf{Return} $k$ models  $\{(\sv_f, C^+, C^-, \gamma)_i\}_{i=1}^k$ \label{ln:retallmodels}
\State  \textbf{else if} $f$ \text{ is not the finest level and } $k=1$
\State \quad \textbf{Return} $(\sv_f, C^+, C^-, \gamma)_1$ \Comment{Return a single model with updated parameters}
\State \textbf{else if } $f$ \text{ is not the finest level and } $k>1$
\NoNumber{\Comment{Return all support vectors from all models and last inherited single parameter set}}
\State  \quad \textbf{Return} $(\cup\{\sv_{f} \text{ from model }i \}_{i=1}^k, \tilde{C}^+, \tilde{C}^-, \tilde{\gamma}$) \label{ln:retunionsv}
\end{algorithmic}
\end{algorithm}

%
%
\noindent {\bf Coarsening} (\textsf{coarsen}-IIS in Alg. \ref{alg:dr})  We start with selecting a random independent set of nodes (or points), $I_0$, using one pass over all nodes (i.e., choose a random node to $I_0$, eliminate it with its neighbors from the graph, and choose the next node). The obtained independent set $I_0$ is added to the set of coarse points. Then, we remove $I_0$ from the graph and repeat the same process to find another independent set $I_1$ which is also added to the set of coarse points. The iterations are repeated until $\sum_k |I_k| \leq Q |\CC^{+}_f|$, where $Q$ is a parameter controlling the size of coarse level space. In our experiments, $Q=0.5$. However, experimenting with different $Q\in [0.4,..,0.6]$ does not affect the quality demonstrating the robustness of this parameter. For too small $Q$, the coarsening might be too fast and, thus, similar to clustering-based coarsening. The process for $\CC^-_f$ is similar.

\noindent {\bf Coarsest level} (line \ref{ln:cst}, Alg. \ref{alg:dr}) At the coarsest level $\rho$, when $|\JJ_{(r)}| \leq M^+ + M^- \ll |\JJ_{(0)}|$, we can apply an exact (or computationally expensive) algorithm for training the coarsest classifier. Typically, $|\JJ_{(\rho)}|$ depends on the available computational resources. However, one can also consider some criteria of separability between $\CC_{(\rho)}^+$, and $\CC_{(\rho)}^-$ \cite{wang2008feature}, i.e., if a fast test exists or some helpful data properties are known. In all our experiments, we used a simple criterion limiting $|\JJ_{(\rho)}|$ to 500. Processing the coarsest level includes an application of NUD \cite{huang2007model} model selection to get high-quality classifiers on the difficult data sets. To this end, we obtained a solution of the coarsest level, namely, $\sv_{(\rho)}$, $C^+_{(\rho)}$, $C^-_{(\rho)}$, and $\gamma_{(\rho)}$.



\noindent {\bf Uncoarsening} Given the solution of coarse level $c$, the primary goal of the uncoarsening is to interpolate and refine this solution for the current fine level $f$. Unlike many other multilevel algorithms, in which the inherited coarse solution contains projected variables only, in our case, we inherit not only $\sv_c$ but also parameters for model selection. This is important because the model selection is an extremely time-consuming component of (W)SVM, and can be prohibitive at fine levels of the hierarchy. However, at the coarse levels, when the problem is much smaller than the original, we can apply much heavier methods for the model selection almost without any loss in the total complexity of the framework.

\begin{algorithm}
\caption{\textsf{uncoarsen}-IIS$(\sv_c)$: uncoarsening at level $f$}
\label{alg:iisunc}
\begin{algorithmic}[1]
\State $(N^+_f, N^-_f) \leftarrow$ Find nearest neighbors  of support vectors $\sv_c$ in $G^+_f$ and $G^-_f$\label{ln:kkn}
\State  $T \gets \sv_c \cup N^+_f \cup N^-_f$ \Comment $T$ is a new training set for refinement\label{ln:initT}
\State {\bf Return} $T$
\end{algorithmic}
\end{algorithm}

\begin{algorithm}
\caption{\textsf{refine}-IIS$(\sv_f, \tilde{C}^+, \tilde{C}^-, \tilde{\gamma})$: refinement at level $f$}
\label{alg:iisref}
\begin{algorithmic}[1]
\If {$|\sv_f| < Q_{t}$}
\State $C^O  \gets (\tilde{C}^+, \tilde{C}^-)$; $\gamma^{O} \gets \tilde{\gamma}$ \label{ln:defcenter}
\State $(\sv_f,C^+_f, C^-_f, \gamma_f) \leftarrow$ train (W)SVM using NUD (or similar technique) initialized with $(C^O, \gamma^{O})$ \label{ln:centernud}
\State {\bf Return} $\sv_f$, $C^+_f$, $C^-_f$, $\gamma_f$

\Else
\State $C^+_f \leftarrow \tilde{C}^+$; $C^-_f \leftarrow \tilde{C}^-$; $\gamma_f \leftarrow \tilde{\gamma}$ \Comment Inherit the coarse parameters \label{ln:inher}
  \State $CL\leftarrow$ partition $\sv_f$ into $K$ (almost) equal size clusters \label{ln:parT}
  \State $\forall k\in CL$ find $P$ nearest opposite-class clusters
  \State $\{(\sv_f, C^+_f, C^-_f, \gamma_f)_i\}_{i=1}^k \leftarrow $ train (W)SVMs on pairs of nearest clusters  with inherited initial parameters $C^+_f$, $C^-_f$, $\gamma_f$, and generate $k$ models \label{ln:trainpairs}
\State  {\bf Return} $k$ models  $\{(\sv_f, C^+_f, C^-_f, \gamma_f)_i\}_{i=1}^k$\label{ln:retref}

\EndIf

\end{algorithmic}
\end{algorithm}

The uncoarsening and refinement are presented in Algorithms  \ref{alg:iisunc} and \ref{alg:iisref}, respectively. After the coarsest level is solved exactly and reinforced by the model selection (line \ref{ln:cst} in Alg. \ref{alg:dr}), the coarse support vectors $\sv_c$ and their nearest neighbors (in our experiments no more than 5) in both classes (i.e., $N^+_f$ and $N^-_f$) initialize the fine level training set $T$ (lines \ref{ln:kkn}-\ref{ln:initT} in Alg. \ref{alg:iisunc}). This completes \textsf{uncoarsen}-IIS (the uncoarsening of $\sv_c$), and $T$ initializes $\sv_f$.

In Alg. \ref{alg:iisref}, the refinement first verifies if $|\sv_f|$ is still small (relatively to the existing computational resources, and the initial size of the data) for applying model selection, i.e., if it is less than a parameter $Q_t$, then we use coarse parameters $\tilde{C}^{+(-)}$, and $\tilde{\gamma}$ as initializers for the current level NUD grid search, and re-train (lines \ref{ln:defcenter}-\ref{ln:centernud} in Alg. \ref{alg:iisref}). Otherwise, the coarse $\tilde{C}^{+(-)}$, and $\tilde{\gamma}$ are inherited in $C^{+(-)}_f$, and $\gamma_f$ (line \ref{ln:inher} in Alg. \ref{alg:iisref}). Then, being large for a direct application of the (W)SVM, $T$ is partitioned into several equal size clusters (using fast solver of balanced $k$-partitioning \cite{bmsss13}), and pairs of nearest opposite clusters are trained (see details in Section \ref{ref_partitioning}). The obtained $K$ models are returned (lines \ref{ln:parT}-\ref{ln:retref} in Alg. \ref{alg:iisref}). If the current level $f$ is finest then we return all models (line \ref{ln:retallmodels} in Alg. \ref{alg:dr}) otherwise a returned union of support vectors and parameter initializations will pass to the next level (see line \ref{ln:retunionsv} in Alg. \ref{alg:dr}). 
\emph{We note that partition-based retraining can be done in parallel, as different pairs of clusters are independent. Moreover, the total complexity of the algorithm does not suffer from reinforcing the partition-based retraining with model selection.}

This coarsening scheme is one of the fastest and easily implementable. While the entire framework (including uncoarsening) is definitely much faster than a regular (W)SVM solver such as LibSVM (which is used in our implementation as a refinement), it is not the fastest among the multilevel SVM frameworks. There is a typical trade-off in discrete multilevel frameworks \cite{cheval-mlpartcompar,SafroRB08}, namely, when the quality of coarsening suffers, the most work is done at the refinement. A similar independent set coarsening approach was used in multilevel dimensionality reduction \cite{fang2010multilevel}. However, in contrast to that coarsening scheme, we found that using only one independent set (including possible maximization of it) does not lead to the best quality of classifiers. Instead, a more gradual coarsening makes the framework much more robust to the changes in the parameters and the shape of data manifold.

\subsection{AMG multilevel framework}
\label{ref_amg_real_point}

The algebraic multigrid (AMG) (W)SVM multilevel framework (\mlsvm-AMG, Alg. \ref{alg:dr}) is inspired by the AMG aggregation solvers for computational optimization problems such as  \cite{amg-sss12,safro-fastresp,KushnirGB06,SafroT11}. Its first version was briefly presented in \cite{sadrfaridpour2016algebraic}. The AMG coarsening generalizes the independent set and clustering \cite{pmlr-v32-hsieha14} based approaches leveraging a high quality coarsening and flexibility of AMG which belongs to the same family of multiscale  learning strategies with the same main phases, namely,  coarsening, coarsest scale learning, and uncoarsening. However, instead of eliminating a subset of the data points, in AMG coarsening, the original problem is gradually restricted to smaller spaces by creating \emph{aggregates of fine data points and their fractions} (which is an important feature of AMG), and turning them  into the data points at coarse levels.  
The main mechanism underlying the coarsening phase is the AMG \cite{mgbooktrott,vlsicad} which successfully helps to identify the interpolation operator for obtaining a fine level solution from the coarse aggregates.
%
In the uncoarsening phase, the solution obtained at the coarsest level (i.e., the support vectors and parameters) is gradually projected back to the finest level by interpolation and further local refinement of support vectors and parameters. A critical difference between AMG approach and the earlier work of Razzaghi et al. \cite{razzaghi2015scalable} is that in AMG approach the coarse level support vectors are not the original data points prolongated from the finest level. Instead, they are centroids of aggregates that contain both full fine-level data points and their fractions.\\

\noindent {\bf Framework initialization} The AMG framework is initialized with $G^{+(-)}_0$ with the edge weights that represent the strength of connectivity between nodes in order to ``simulate'' the following  interpolation scheme applied at the uncoarsening, in which strongly coupled nodes can interpolate solution to each other. In the classifier learning problems, this is expressed as a similarity measure between points. We define a distance function between nodes (or corresponding data points) as an inverse of the Euclidean distance. More advanced distance measure approaches such as \cite{brannick2006energy,ChenS11} are often essential in similar multilevel frameworks.

\noindent {\bf Coarsening Phase }(see Algorithm \ref{alg:coarsening}, $\textsf{coarsen}$-AMG)
We describe the two-level process of obtaining the coarse level training set $\CC^+_c$ with corresponding $G^+_c$ given the current fine level $G^+_f$ and its training set (e.g., the transition from level $f$ to $c$). The majority class is coarsened similarly. 

The process is started with selecting seed nodes that will serve as centers of coarse level nodes, i.e., the aggregates at level $f$. Coarse nodes will correspond to the coarse data points at level $c$. Structurally, each aggregate must include one full seed $f$-level point, and possibly several other $f$-level points and their fractions. Intuitively, it is equivalent to grouping points in $\CC^+_f$ into many small subsets allowing intersections, where each subset of nodes  corresponds to a coarse point at level $c$. 
During the aggregation process, most coarse points will correspond to aggregates of size greater than 1 (because, throughout the hierarchy, they accumulate many fine points and their fractions), so we introduce the notion of a \emph{volume} $v_i\in \RR_+$ for all $i\in \CC^+_f$ to reflect the importance of a point or its capacity that includes finest-level aggregated points and their fractions. We also introduce the edge weighting function $w:E^+_f\rightarrow \RR_{\geq 0}$ to reflect the strength of connectivity and similarity between nodes.

In Algorithm \ref{alg:coarsening}, we show the details of AMG coarsening.
In the first step (line \ref{ln:fvol}), we compute the future-volumes $\vartheta_i$ for all $i\in \CC^+_f$ to determine the order in which $f$-level points will be tested for declaring them as seeds, namely,
\begin{equation}
\label{eqn:futurevolume}
\vartheta_i = v_i + {\sum\limits_{j\in \Gamma_i\cap \CC^+_f} v_j \cdot {w_{ji} \over \sum\limits_{k\in \Gamma_j \cap  \CC^+_f} w_{jk}}},
\end{equation}
where $\Gamma_i$ is the neighborhood of node $i$ in $G^+_f$. The future-volume $\vartheta_i$ is defined as a measure (that is often used in multilevel frameworks \cite{SafroRB08}) of how much an aggregate 
seeded by a point $i$ may potentially grow at the next level $c$. This is computed in linear time.

We assume that in the finest level, all volumes are ones. We start with selecting a dominating set of seed nodes $S\subset \CC^+_f$ to initialize aggregates. Nodes that are not selected to $S$ remain in $F$ such that $\CC^+_f = F\cup S$. Initially, the set $F$ is set to be $\CC^+_f$, and $S=\emptyset$  since no seeds have been selected. After that, points with $\vartheta_i > \eta \cdot \overline{\vartheta}$, i.e., those that are exceptionally larger than the average future volume are transferred to $S$ as the most ``representative'' points (line \ref{ln:largevol}). 
%
Then, all points in $F$ are accessed in the decreasing order of $\vartheta_i$ updating $S$ iteratively (lines \ref{ln:iterseeds1}-\ref{ln:iterseeds2}), namely, if with the current $S$, and $F$, for point $i\in F$, $\sum_{j \in S} w_{ij} / \sum_{j \in \CC^+_f} w_{ij}$ is less than or equal to some threshold $Q$\footnote{Similar parameter $Q$ that controls the speed of coarsening appears in \textsf{coarsen}-IIS.}, i.e., the point is not strongly coupled to already selected points in $S$, then $i$ is moved from $F$ to $S$.

%
%
%
The points with large future-volumes usually have a better chance to serve as seeds and become centers of future coarse points. 
Selecting too few seeds (and then coarse level points)  causes ``overcompressed'' coarser level which typically leads to the classification quality drop. Therefore, in order to keep sufficiently many points at the coarse level, the parameter $Q$ is set to 0.4-0.6. It has been observed that in most AMG algorithms, $Q > 0.6$ is not required (however, it depends on the type and goals of aggregation). In our experiments $Q=0.5$, and $\eta = 2$. Other similar values do not significantly change the results.
\begin{algorithm}
  \caption{\textsf{coarsen}-AMG($\CC_f^+, G_f^+$): AMG coarsening}\label{alg:coarsening}
  \begin{algorithmic}[1]
  \State $S\gets \emptyset, F\gets \CC^+_f$ \Comment{start select seeds for coarse nodes} 
      \State {\bf Calculate} using Eq. (\ref{eqn:futurevolume}) $\forall i\in F$ $\vartheta_i$, and the average $\bar{\vartheta}$\label{ln:fvol}
      \State $S \gets$ nodes with $\vartheta_i > \eta \cdot \overline{\vartheta}$ \label{ln:largevol}
      \State $F \gets V_f \setminus S$
      \State {\bf Recompute} $\vartheta_i$ $\forall i \in F$
      \State Sort $F$ in descending order of $\vartheta$
      \For{$i \in F$} \label{ln:iterseeds1}
	      \If{$\left({\sum\limits_{j \in S} w_{ij}} / {\sum\limits_{j \in \JJ^+_f} w_{ij}}\right) \leq Q$} \label{ln:paramQ}
		      \State move $i$ from $F$ to $S$
	      \EndIf
      \EndFor \Comment{end select seeds for coarse nodes}  \label{ln:iterseeds2}
      \State Build interpolation matrix $P$ according to Eq. (\ref{eq:interp})
      \State Build coarse graph $G^+_c$ with edge weights using Eq. (\ref{eq:edgew})
      \State Define volumes of coarse points using Eq. (\ref{eq:cvol})
      \State Compute coarse points $\CC_c^+$ using Eq. (\ref{eq:pointagg})
      \State {\bf Return} $(\CC_c^+, G_c^+)$
  \end{algorithmic}
\end{algorithm}

When the set $S$ is selected, we compute the AMG interpolation matrix $P\in \RR^{|\CC^+_f|\times |S|}$ that is defined as
\begin{equation}\label{eq:interp}
P_{ij} =
\left \{
\begin{tabular}{cc}
$ {w_{ij}} / {\sum\limits_{k \in \Gamma_i} w_{ik}} $ & if $i \in F$, $j \in \Gamma_i$ \\
1 & if $i \in S$, $j=I(i)$\\
0 & otherwise
\end{tabular}
\right \},
\end{equation}
where $\Gamma_i = \{j \in S \mid ij\in E^+_f\}$ is the set of $i$th seed neighbors, and $I(i)$ denotes the index of a coarse point at level $c$ that corresponds to the fine level aggregate around seed $i\in S$. Typically, in AMG methods, the number of non-zeros in each row is limited by the parameter called the interpolation order or caliber \cite{vlsicad} (see further discussion about $r$ and Table \ref{tbl:interpolation_order_effect}). This parameter, $r$, controls the complexity of a coarse-scale system (the number of non-zero elements in the matrix of coarse $k$-NN graph). It limits the number of fractions a fine point can be divided into (and thus attached to the coarse points). If a row in $P$
contains too many non-zero elements then it is likely to increase the number of non-zeros in the coarse graph matrix. In multigrid methods, this number is usually controlled by different approaches that measure the strength of connectivity (or importance) between fine and coarse variables (see discussion and implementation in \cite{safro:relaxml}). 


Using the matrix $P$, the aggregated data points and volumes for the coarse level are calculated.
The edge between points $p=I(i)$ and $q=I(j)$ is assigned with weight
\begin{equation}\label{eq:edgew}
w_{pq} = \sum\nolimits_{k \ne l} P_{ki} \cdot w_{kl} \cdot P_{lj}.
\end{equation}
The volume for the aggregate $I(i)$ in the coarse graph is computed by
\begin{equation}\label{eq:cvol}
\sum\nolimits_{j} v_j P_{ji},
\end{equation}
i.e., the total volume of all points is preserved at all levels during the coarsening. The coarse point $q\in \CC^+_c$ seeded by $i=I^{-1}(q)\in \CC^+_f$ is represented by 
\begin{equation}\label{eq:pointagg}
\sum_{j\in \mathcal{A}_i} P_{j,q} \cdot j,
\end{equation}
where $\mathcal{A}_i$ is a set of fine points in aggregate $i$. This set is extracted from the column of $P$ that corresponds to aggregate $i$ by considering rows $j$ with non-zero values.
%
%

The stopping criteria for the coarsening depends on the available computational resources that can be used in order to train the classifier at the coarsest level. In our experiments, the coarsening stops when the size is less than a threshold (typically, 500 points) that ensures a fast performance of the \\LibSVM dual solver.\\

\noindent {\bf Uncoarsening} (see Algorithm \ref{alg:coarsening}, $\textsf{uncoarsen}$-AMG)
The uncoarsening of AMG multilevel framework is similar to that of the \mlsvm-IIS. The main difference is in lines \ref{ln:kkn}-\ref{ln:initT} in Alg. \ref{alg:iisunc}. Instead of defining the training set for the refinement at level $f$ as 
\[
T \gets \sv_c \cup N^+_f \cup N^-_f,
\]
all coarse support vectors are uncoarsened by adding to $T$ all elements of the corresponding aggregates, namely,

\begin{align}\label{eq:amg-disagg} 
T \gets \emptyset; ~ &\quad  \forall p\in \sv_c \quad \forall j\in \mathcal{A}_p \quad T \gets T \cup j.
\end{align}
The rule in (\ref{eq:amg-disagg}) means the following: 1) take all $c$-level support vectors $p$, 2) find all $f$-level points that are aggregated in $c$-level support vectors, and 3) add them to $T$. 
The basic refinement, \textsf{refine}-AMG, is similar to \textsf{refine}-IIS.

\subsection{Complexity of multilevel framework}\label{sec:complexity}

The complexity of MAF for (W)SVM consists of three parts, namely, generating approximated $k$-NN graphs of both classes, coarsening and uncoarsening. 
The complexity of generating approximate $k$-NN graphs is based on FLANN library implementation  \cite{muja_flann_2009,muja2014scalable} that was used in our experiments. It includes construction of a k-means tree that is leveraged to search for approximate nearest neighbors. 
The overall complexity of FLANN is $\OO(|\JJ|\cdot d\cdot (\log n'/ \log K))$ where $d$ is the data dimensionality, $n'$ is the number of inner nodes the k-means tree, and $K$ is the number of clusters or branching factor for the k-means.
When we compare the running time of 1 V-cycle of our solver and that of parallelized FLANN preprocessing, we observe that FLANN does not significantly increases the running time of the entire framework when we parametrize it to find 10 nearest neighbors. 

In the coarsening phase, we need to consider the complexity of coarsening the approximated $k$-NN graphs of $\CC^+$ and $\CC^-$ including aggregation of the data points. The complexity of coarsening is similar to that of AMG applied on graph $G=(V,E)$ which is proportional to $|V| + |E|$, where $|E|\approx k|V|$, where $k$ is the number of nearest neighbors. In our experiments, we found that no data set requires $k>10$ to improve the quality of classification. Because we do not anticipate to obtain exceptionally high-degree nodes during the coarsening, we also do not expect to observe very fast increasing density of nonzero features (nnz) in data points. Thus, we bound the complexity of coarsening with $\OO(\text{nnz}(\JJ))$ (or $\OO(|\JJ|)$ for low dimensional data) without having hidden coefficients, in practice.

The complexity of the uncoarsening mostly depends on that of the underlying QP solver (call it QPS, such as LibSVM) applied at the refinement stage. Another factor that affects the complexity is the number of support vectors found at each scale which is typically significantly smaller than the number of data points. Typically, the complexity will be approximately $\OO(nnz(\JJ)) + \OO(\text{QPS}(p\text{ points})) \cdot |\text{support vectors}|/p$, where $p$ is the number of parts, the set of support vectors is split to if partitioning is applied. Typically, if the application does not include very dense data, the component $\OO(nnz(\JJ))$ is much smaller than $\OO(|\JJ|\cdot d)$. Overall, the complexity of the entire framework is linear in the number of data points.

The computational time obtained in our experiments and the amount of work per unit is presented in Section \ref{sec:compresults}. In particular, in Table \ref{tbl:complex} we demonstrate the computational time per data point and per feature value. In particular, in Figure \ref{fig:scal}, we present the change in running time while training the model with  increasingly larger parts of the dataset.

\subsection{Engineering multilevel framework}
\label{alg_amg_agg_point}
The AMG framework generalizes many multilevel approaches by allowing a ``soft'' weighted aggregation of points (and their fractions) in contrast to the ``strict'' clustering \cite{pmlr-v32-hsieha14} and subset based aggregations such as our \mlsvm-IIS \cite{razzaghi2015scalable}.
%
%
In this section we describe a variety of improvements we experimented with to further boost the quality of the multilevel classification framework, and improve the performance of both the training and validation processes in terms of the quality and running time. All of them are applicable in both ``strict'' or ``soft'' coarsening schemes.


\subsubsection{Imbalanced classification} 

One of the major advantages of the proposed coarsening scheme is its natural ability to cope with the imbalanced data in addition to the cost-sensitive and weighted models solved in the refinement. When the coarsening is performed on both classes simultaneously, and in a small class the number of points reaches an allowed minimum, this level is simply copied throughout the rest of levels required to coarsen the big class. Since the number of points at the coarsest level is small, this does not affect the overall complexity of the framework.
Therefore, the numbers of points in both classes are within the same range at the coarsest level regardless of how imbalanced they were at the fine levels.
Such training on the balanced data mitigates the imbalance effects and improves the performance quality of trained models.

\subsubsection{Coarse level density problem}
\label{alg_amg_agg_point_coarsening}

Both \mlsvm-IIS and \mlsvm-AMG do not change the dimensionality during the coarsening which potentially may turn into a significant computational bottleneck for a large-scale data. In many applications, a high-dimensional data is sparse, and, thus, even if the number of points is large, the space requirements are still not prohibitive. Examples include text tf-idf data and categorical features that are converted into a binary representation. While the \mlsvm-IIS coarsening selects \emph{original} (i.e., sparse) points for the coarse levels, the \mlsvm-AMG aggregates points using a linear combination such as in Eq. \ref{eq:pointagg}. Even when the original $j\in \mathcal{A}_i$ are sparse, the points at coarse levels may eventually become much denser than the original points.

The second type of coarse level density is related to the aggregation itself. When $f$-level data points are divided into several parts to join several aggregates, the number of edges in coarse graphs is increasing when it is generated by $L_c\gets P^T L_f P$ (subject to $L_c$ diagonal entry correction). Finest level graphs that contain high-degree nodes have a good chance to generate very dense graphs at the coarse levels if their density is not controlled. This can potentially affect the performance of the framework. 

The coarse level density problem is typical to most AMG and AMG-inspired approaches. We control it by filtering weak edges and using the order of interpolation in aggregation. 
The weak edges not only increase the density of coarse levels but also may affect the quality of the training process during the refinement. 
In \mlsvm-AMG framework, we eliminate weak edges between $i$ and $j$ if 
$w_{ij} < \theta \cdot \text{avg}_{ki}\{w_{ki}\}$  
and $w_{ij} < \theta \cdot  \text{avg}_{kj}\{w_{kj}\}$ where $\text{avg}\{\cdot\}$ is the average of corresponding adjacent edge weights.
We experimented with different values of $\theta$ between $0.001$ and $0.005$  which was typically a robust parameter that does not require much attention.

The order of interpolation, $r$, is the number of nonzeros allowed per row in $P$. A single nonzero $j$th entry in row $i$, $P_{ij}=1$, means that a fine point $i$ fully belongs to aggregate $j$ which leads to creation of small clusters of fine points without splitting them. Typically, in AMG methods, increasing $r$ improves the quality of solvers making them, however, slower. We experiment with different values of $r$ and conclude that high interpolation orders such as 2 and 4 perform better than 1. In the same time, we observed that there is no practical need to increase it more (see further discussion and example in Sec. \ref{sec:r1amg}).

\subsubsection{Validation for model selection}
\label{ref_validation_data}
The problem of finding optimal parameters (i.e., the model selection) is important for achieving a better quality on many data sets. Typically, this component is computationally expensive because repetitive training is required for different choices of parameters. A validation data is then required to choose the best trained model.
A performance of model selection techniques is affected by the quality and size of the validation data. (We note that the test data for which the computational results are presented remains completely isolated from any training and validation.) 

The problem of a validation set choice requires a special attention in multilevel frameworks because the models at the coarse levels should not necessarily be validated on the corresponding coarse data. As such, we propose different approaches to find the most suitable types of validation data.
%
We developed the following approaches to  choose validation set for multilevel frameworks, namely, \emph{coarse sampling} (CS), \emph{coarse cross k-fold} (CCkF),  \emph{finest full} (FF), and \emph{fine sampling} (FS).\\
%
%
%


\noindent {\bf CS:} The data in $\JJ^+_{(i)}$ and $\JJ^-_{(i)}$ is sampled and one part of it (in our experiments 10\% or 20\%) is selected for a validation set for model selection. In other words, the validation is performed on the data at the same level. This approach is extremely fast on the data in which the coarsening is anticipated to be uniform without generating a variability in the density of aggregation in different parts of the data. Typically, its quality is acceptable on homogeneous data. However, qualitatively, this approach may suffer from a small size of the validation data in comparison to the size of test data.\\
%
%
%
\noindent {\bf CCkF:} In this method we apply a complete k-fold cross validation at all levels using the coarse data from the same level. The disadvantage of this method is that it is more time consuming but the quality is typically improved. During the k-fold cross validation, all data is covered. With this method, the performance measures are improved in comparison to the CS but the quality of the finest level can degrade because of potential overfitting at the coarse levels.

\noindent {\bf FF:} This method exploits a multilevel framework by combining a coarse training set $\JJ^{+(-)}_c$ with a validation set that is a whole finest level training set $\JJ^{+(-)}_{(0)}$. The idea behind this approach is to choose the best model which increases a required performance measure (such as accuracy, and G-mean) of coarse aggregates with respect to the original data rather than the aggregates. This significantly increases the quality of final models. However, this method is time consuming on very large data sets as all original points participate in validation.

\noindent {\bf FS:} This method resolves the complexity of FF by sampling  $\JJ^{+(-)}_{(0)}$ to serve as a validation set at the coarse levels. The size of sampling should depend on computational resources. However, we note that we have not observed any drop in quality if it is more than 10\% of the  $\JJ^{+(-)}_{(0)}$. Both FF and FS exhibit the best performance measures.

\subsubsection{Underlying solver}
At all iterations of the refinement and at the coarsest level we used LibSVM  \cite{chang2011libsvm} as an underlying solver by applying it on the small subsets of data (see lines \ref{ln:centernud} and \ref{ln:trainpairs} in Alg. \ref{alg:iisref}). Depending on the objective Eq. (\ref{eqn:SVM}) or (\ref{eqn:WSVM}), SVM or WSVM solvers are applied. In this paper we report the results of WSVM in which the objective Eq. (\ref{eqn:WSVM}) is given by
\begin{align}\label{eqn:WSVM2}
  \textsf{minimize} ~~~~~ & \frac{1}{2}\lVert w \rVert^2 + C\big( {W^+}\sum_{i=1}^{n^+}  \xi_i^+ + {W^-}\sum_{j=1}^{n^-}  \xi_j^- \big) , 
\end{align}
where the optimal $C$ and $\gamma$ are fitted using model selection, and the class importance coefficients are $W^+$ and $W^-$. While for the single-level WSVM, the typical class importance weighting scheme is 
\begin{equation}
W^+= \frac{1}{ |\JJ^+| } , \quad W^-= \frac{1}{ |\JJ^-| },
\end{equation}
in MAF, the aggregated points in each class have different importance due to the different accumulated volume of finer points. 
The aggregated points which represent more fine points are more important than aggregated points which represent small number of fine points. 
Therefore, the MAF approach for calculating the class weights is based on sum of the volumes in each class, i.e.,
\begin{equation}\label{eq:clweight}
W^+= \frac{1}{ \sum\limits_{i \in \JJ^+}^{} v_i } , \quad W^-= \frac{1}{ \sum\limits_{i \in \JJ^-}^{} v_i }.
\end{equation}

This method, however, ignores the importance of each point in its class. We find that the most successful penalty scheme is the one that is personalized per point, and adapt (\ref{eq:clweight}) to be
\begin{equation}
\forall i\in \JJ^{+(-)} \quad W_i= W^{+(-)}\frac{v_i}{ \sum\limits_{j \in \JJ^{+(-)}}^{} v_j }.
\end{equation}
In other words, we consider the relative volume of the point in its class but also multiply it by an inverse of the total volume of the class which gives more weight to a small class. This helps to improve the correctness of a small class classification.
\subsubsection{Expanding training set in refinement}

Typically, in many applications, the number of support vectors is much smaller than $|\JJ|$. This observation allows some freedom in exploring the space around support vectors inherited from coarse levels. This can be done  by adding more points to the refinement training set in attempt to improve the quality of a hyperplane. We describe several possible strategies that one can follow in designing a multilevel (W)SVM framework.

\noindent {\bf Full disaggregation:} This is a basic method presented in (\ref{eq:amg-disagg}) in which all aggregates of coarse support vectors contribute all their elements to the training set. It is a default method in \mlsvm-AMG. \\
\noindent {\bf $k$-distant disaggregation:} In some cases, the quality can be improved by adding to $T$ the $k$-distant neighbors of aggregate elements. In other words, after (\ref{eq:amg-disagg}), we apply
\begin{equation}
\forall p\in T \quad T\gets T\cup N^{+(-)}_f(p),
\end{equation}
where $N^{+(-)}_f(p)$ is a set of neighbors of $p$ in $G^{+(-)}_f$ depending on the class of $p$. Similarly, one can add points within distance $k$ from the aggregates of inherited support vectors. Clearly, this can only improve the quality. However, the refinement training is expected to be increasingly slower especially when the $G^{+(-)}_f$ contains high-degree nodes. Even if the finest level graph nodes are of a small degree, this could happen at the coarse levels if a proper edge filtering and limiting interpolation order are not applied. In very rare cases, we observed a need for adding distance 2 neighbors.

\noindent {\bf Sampling aggregates:} In some cases, the coarse level aggregates may become very dense which does not improve the quality of refinement training. Instead, it may affect the running time. One way to avoid of unnecessary complexity is to sample the elements of aggregates. A better way to sample them than a random sampling (after adding the seed) is to order them by the interpolation weights $P_{ij}$. The ascending order which gives a preference to the fine points that are split across more than one aggregate was the most successful option in our experiments. Fine non-seed points whose $P_{ij}=1$ are likely to have high similarity with the seeds which does not improve the quality of the support vectors.

\subsubsection{Partitioning in the refinement}
\label{ref_partitioning}
When the number of uncoarsened support vectors at level $f$ is too big for the available computational resources, multiple small-size models are trained and either validated or used as a final output. Small-size models are required for applying model selection in a reasonable computational time. For this purpose we partition the current level training sets $\CC^{+(-)}_f$ (see lines \ref{ln:parT}-\ref{ln:retref} in Alg. \ref{alg:iisref}) into   $k$ parts of approximately equal size using fast \emph{graph partitioning} solvers \cite{bmsss13}. Note that applying similar \emph{graph clustering} strategies may lead to highly imbalanced parts which will make the whole process of refinement acceleration useless.

In both \mlsvm-AMG and \mlsvm-IIS, we leverage the graphs of both classes $G^+_f$ and $G^-_f$ with the inverses of Euclidean distance between nodes playing the role of edge weights. After both graphs are partitioned, two sets of approximately equal size partitions, $\Pi^+_f$ and $\Pi^-_f$ are created. For each part $\pi_i\in \Pi^{+}_f \cup \Pi^{-}_f$ we compute its centroid $c_i$ in order to estimate the nearest parts of opposite classes and train multiple models by pairs of parts.

The training by pairs of parts works as follows. For each $c_i$ we find the nearest $c_j$ such that $i$ and $j$ are in different classes and evaluate at most $|\Pi^+_f| + |\Pi^-_f|$ models for different choices of $(\pi_i,\pi_j)$ pairs (without repetitions which often appear in practice making the process fast). The set of all generated models is denoted by $\MM_f$. We note that the training of such pairs is independent and can be easily parallelized. 

There are multiple ways one can test (or validate) a point using all models ``voting''. The simplest strategy which performs well on many data sets is a majority voting. However, the most successful way to generate a prediction was a voting by the relative distance from the test point $t$ to the weighted center of the segment connecting $c_i$ and $c_j$, namely, 
\begin{equation}
x_{ij} = \frac{c_i\sum_{q\in \pi_i}v_q + c_j\sum_{q\in \pi_j}v_q}{\sum_{q\in \pi_i\cup \pi_j}v_q}, 
\end{equation}
where $v_q$ is the volume of point $q$. For all pairs of nearest parts $i$ and $j$, the label of $t$ is computed as
\begin{equation}\label{eq:multmodpred}
\text{sign}\big(\frac{\sum_{ij\in \MM_f} l_{ij}(t) d^{-1}(t, x_{ij})}{\sum_{ij\in \MM_f} d^{-1}(t, x_{ij})}\big),
\end{equation}
where $l_{ij}(t)$ is a label of $ij$ model for point $t$, and $d(\cdot, \cdot)$ is a distance function between two points. We experimented with several distance functions to express the proximity of parts (i.e., the way we choose pairs $(\pi_i,\pi_j$)) and $d(\cdot, \cdot)$, namely, Euclidean, exponential, and Manhattan. The quality of final models obtained using Euclidean distance was the highest.

If the partitioning refinement is applied at the finest level then Algorithm \ref{alg:dr} outputs all generated finest level models, and the prediction works according to Eq. (\ref{eq:multmodpred}). Otherwise, if the partitioning refinement occurs in the middle levels then the next finer level will receive a union of all support vectors from the models (line \ref{ln:retunionsv} in Alg. \ref{alg:dr}) and model parameters inherited from last level in which a single model was trained. We note that it often might be the case that a partitioning refinement generates models with relatively small total number of support vectors such that at the next finer level, their union can be considered as an input to train a single model.

\subsubsection{Model Selection}
\label{ref_model_selection}

The MAF allows a flexible design for model selection techniques such as various types of parameter grid search \cite{chapelle2002choosing},  NUD \cite{huang2007model} that we use in our computational experiments, and other search approaches \cite{lin2008parameter,bao2013pso,zhang2010aco}. A mechanism that typically works behind most of such search techniques evaluates different combinations of parameters (such as $C^+$, $C^-$, and $\gamma$) and chooses the one that exhibits the best performance measure. Besides the general applicability of model selection because the number of inherited and disaggregated support vectors (in the uncoarsening of \mlsvm-IIS and \mlsvm-AMG) is typically smaller than that of the corresponding training set, the MAF has the following advantages.

\noindent {\bf Fast parameter search:} In many cases, there is no need to test all combinations of the parameters. The inherited $c$-level parameters can serve as a center point for their refinement only. For example, NUD suggests two-stage search strategy. In the first stage a wide range of parameters is considered. In the second stage, the best combination from the first stage is locally refined using a smaller search range. In MAF, we do not need to apply the first stage as we only refine the inherited $c$-level parameters. Other grid search methods can be adjusted in a similar way.

\noindent {\bf Selecting suitable performance measures for the best model:} In MAF, a criterion for choosing the best model throughout the hierarchy is more influential than that at the finest level in non-MAF frameworks. Moreover, these criteria can be different at different levels.  For example, when one focuses on highly imbalanced sets, a criteria such as the best G-mean could be more beneficial than the accuracy. We found that introducing 2-level criteria for imbalanced sets such as (a) choose the best G-mean, and (b) among the combinations with the best G-mean choose the best sensitivity, performs particularly good if applied at the coarse levels when the tie breaker may be often required.

\subsubsection{Models at different levels of coarseness}\label{sec:mod-scales}

Over- and under-fitting are among the key problems of model selection and classifiers, in general. The MAF successfully helps to tackle them. 
Throughout the hierarchy, we solve (W)SVM models at different levels of coarseness. Intuitively, the coarsening procedure gradually creates generalized (or summarized) representations of the finest level data which results in generalized coarse hyperplanes which can also be used as \emph{final solutions}. Indeed, at the finest level, rich data can easily lead to over-fitted models, a phenomenon frequently observed in practice \cite{dietterich1995overfitting}. In the same time, over-compressed data representation may lead to an under-fitted model because no fine details are considered. 
In a multilevel framework, one can use models from multiple levels of coarseness because the most correct validation is done against the fine level data in any case. Our experiments confirm that more than half of the best models are obtained from the coarse (but not coarsest) and middle levels which typically prevents over- and under-fitting.

If the best validation was obtained at the middle level and at this level the framework generated multiple models using partitioning refinement (see Section \ref{ref_partitioning}) then these multiple models will be the output of Alg. \ref{alg:dr} and the prediction will work according to Eq. (\ref{eq:multmodpred}). In general, if the best models were produced by the  finest and middle levels, we recommend to use the middle level model to avoid potential over-fitting. This recommendation is based on the observation that same quality models can be generated by different hyperplanes but finest models may contain a large number of support vectors that can lead to over-fitting. However, it is a general thought that requires further exploration. In our experiments, no additional parameters or conditions are introduced to choose the final model. We simply choose the best model among those generated at different levels.

\section{Computational Results}\label{sec:compresults}

We compare our algorithms in terms of classification quality and computational performance to the state-of-the-art sequential SVM algorithms LibSVM, DC-SVM, and fast Ensemble SVM. The DC-SVM is a most recent, fast, hierarchical approach that outperforms other hierarchical methods which was the reason to choose it for comparison. The classification quality is evaluated using the following performance measures: sensitivity (SN), specificity (SP), geometric mean (G-mean), and accuracy (ACC), Precision (PPV), and F1, namely,
\begin{equation*}
\textrm{SN}=\frac{TP}{TP+FN}, \ \ \textrm{SP}=\frac{TN}{TN+FP}, \ \ \textrm{G-mean}=\sqrt{\textrm{SP}\cdot \textrm{SN}}, 
\end{equation*}
\begin{equation*}
\textrm{ACC}=\frac{TP+TN}{FP+TN+TP+FN},  \ \ \textrm{Precision (PPV)}=\frac{TP}{TP+FP}, 
\end{equation*}
where $TN$, $TP$, $FP$, and $FN$ correspond to the numbers of true negative, true positive, false positive, and false negative points. Our main metric for comparison is G-mean which measures the balance between classification quality  on both the majority and minority classes. This metric is illuminating for imbalanced classification as a low G-mean is an indication of low-quality classification of the positive data points even if the negative points classification is of high quality. This measure indicates over-fitting of the negative class and under-fitting of the positive class, a critical problem in imbalanced datasets. 

In all experiments the data is normalized using z-score. Each experimental result in the following tables represents an average over 100 executions of the same type with different random seeds. 
The computational time reported in all experiments contains generating the $k$-NN graph. The computational time is reported in seconds unless it is explicitly mentioned otherwise.

%
In each class, a part of the data is assigned to be the test data using $k$-fold cross validation. We experimented with $k$=5 and 10 (no significant difference was observed). 
The experiments are repeated $k$ times to cover all the data as test data. The data randomly shuffled for each $k$-fold cross validation.
The presented results are the averages of performance measures for all $k$  folds. 
Data points which are not in the test data are used as the training data in $\JJ^{+(-)}$. 
The test data is never used for any training or validation purposes. 
The Metis library \cite{KarypisKumar98metis} is used for graph partitioning during the refinement phase.
%
We present the details about data sets in Table \ref{tbl_dataset_info}. The imbalance of datasets is denoted by $\epsilon$.

\begin{table}[htb]
\centering
\caption{Benchmark data sets.}
\label{tbl_dataset_info}
\scalebox{.8}{
    \begin{tabular}{lccccccc}         
	\hline
		Dataset	&	$\epsilon$	&	$n_f$ &	$|\JJ|$	&	$|\CC^+|$	&	$|\CC^-|$	\\ \hline
		Advertisement	  &	0.86	&	1558    &	3279	&	459	    &	2820	\\
		Buzz              &	0.80	&	77	    &	140707	&	27775	&	112932	\\
		Clean (Musk)	  &	0.85	&	166	    &	6598	&	1017	&	5581	\\
		Cod-rna	          &	0.67	&	8	    &	59535	&	19845	&	39690	\\
		EEG Eye State	  &	0.55	&	14	    &	14980	&	6723	&	8257	\\
		Forest (Class 5)  & 0.98    &   54      &   581012  &   9493    &   571519  \\
		Hypothyroid       & 0.94    &   21      &   3919    &   240     &   3679    \\
		ISOLET            & 0.96    &   617     &   6238    &   240     &   5998    \\
		Letter  	      &	0.96	&	16	    &	20000	&	734	    &	19266	\\
		Nursery	          &	0.67	&	8	    &	12960	&	4320	&	8640	\\
		Protein homology  & 0.99    &   74      &   145751  &   1296    &   144455  \\
		Ringnorm	      &	0.50	&	20	    &	7400	&	3664	&	3736	\\
		Twonorm	          &	0.50	&	20	    &	7400	&	3703	&	3697	\\ 	\hline
	\end{tabular}
	}
\end{table}

The Forest data set \cite{UCI} has 7 classes and different classes are reported in the literature (typically, not the difficult ones). Class 5 is used in our experiments as the most difficult and  highly imbalanced.
We report our results on other classes which are listed in Table \ref{table_forest_classes_info} for convenient comparison with other methods. 

\begin{table}[H]
\centering
\caption{The Forest data set classes with $n_f = 54$ and $|\JJ| = 581012$}
\label{table_forest_classes_info}
\scalebox{.8}{
    \begin{tabular}{cccccc}         

        \hline
Class No	& $\epsilon$	& 	$|\CC^+|$	&	$|\CC^-|$	\\ \hline
Class 1     &	0.64	 &	211840  &	369172    \\
Class 2     &	0.51	 &	283301	&	297711    \\
Class 3     &	0.94	&	35754	&	545258    \\
Class 4     &	1.00	&	2747	&	578265    \\
Class 5     &	0.98	&	9493    &	571519    \\
Class 6     &	0.97	&	17367	&	563645    \\
Class 7     &	0.96	&	20510	&	560502    \\
\hline
\end{tabular}
}
\end{table}
\subsection{\mlsvm-IIS results} \label{SecIte} 

The performance measures of single- (LibSVM) and multi-level (W)SVMs are computed and compared in Table \ref{table4}. In our earlier work \cite{razzaghi2015scalable}, it has been shown in that the multilevel (W)SVM produces similar results compared to the single-level (W)SVM, but it is much faster (see Table \ref{table5}). All experiments on all data sets have been executed on a single machine Intel Core i7-4790, 3.60GHz, and 16 GB RAM. The framework ran in sequential mode with no parallelization using Ubuntu 14.04.5 LTS, Matlab 2012a, Metis 5.0.2, and FLANN 1.8.4.

\begin{table}[htb]
\centering
\caption{Quality comparison using performance measures for multi- and single-level of (W)SVM. Each cell contains an average over 100 executions including model selection for each of them.   Column ``Depth'' shows the number of levels. The best results are highlighted in bold font.} 
\label{table4}
 \scalebox{.8}{
    \begin{tabular}{clccccc|cccc}                  
        \hline
 & & \multicolumn{5}{c|}{Multilevel} & \multicolumn{4}{c}{Single-level}\\ \cline{3-5}   \cline{6-11}
& Dataset	&	ACC	&	SN	&	SP	&	G-mean	&	Depth&	ACC	&	SN	&	SP	&	G-mean		\\\hline
\parbox[t]{2mm}{\multirow{11}{*}{\rotatebox[origin=c]{90}{SVM}}} 
	& Advertisement     & 0.94 & 0.97 &	0.79 & \textbf{0.87}   & 7	& 0.92	& 0.99  & 0.45  & 0.67          \\
	& Buzz              & 0.94 & 0.96 &	0.85 & \textbf{0.90}   & 14	& 0.97	& 0.99	& 0.81  & 0.89          \\
	& Clean (Musk)      & 1.00 & 1.00 & 0.99 & \textbf{0.99}   & 5	& 1.00 	& 1.00  & 0.98  & \textbf{0.99} \\
	& Cod-rna	    	& 0.95 & 0.93 &	0.97 & 0.95            & 9	& 0.96 	& 0.96  & 0.95 	& \textbf{0.96} \\
	& EEG Eye State     & 0.83 & 0.82 & 0.88 & 0.85	           & 6  & 0.88 	& 0.90  & 0.86  & \textbf{0.88} \\
	& Forest (Class 5)  & 0.93 & 0.93 & 0.90 & 0.91            & 33 & 1.00  & 1.00  & 0.86  & \textbf{0.92} \\
	& Hypothyroid	    & 0.98 & 0.98 & 0.74 & \textbf{0.85}   & 4	& 0.99	& 1.00  & 0.71  & 0.83          \\
	& ISOLET            & 0.99 & 1.00 & 0.83 & \textbf{0.92}   & 11	& 0.99 	& 1.00  & 0.85  & \textbf{0.92} \\
	& Letter	    	& 0.98 & 0.99 &	0.95 & 0.97	           & 8	& 1.00 	& 1.00 	& 0.97  & \textbf{0.98} \\
	& Nursery	    	& 1.00 & 0.99 &	0.98 & 0.99 	       & 10	& 1.00 	& 1.00  & 1.00  & \textbf{1.00} \\
	& Protein homology  & 1.00 & 1.00 &	0.72 & 0.85	           & 18 & 1.00  & 1.00  & 0.80  & \textbf{0.89}	\\
	& Ringnorm          & 0.98 & 0.98 & 0.99 & \textbf{0.98}   & 6	& 0.98 	& 0.99 	& 0.98  & \textbf{0.98} \\
	& Twonorm	        & 0.97 & 0.98 &	0.97 & 0.97  		   & 6  & 0.98	& 0.98  & 0.99  & \textbf{0.98} \\\hline
\parbox[t]{2mm}{\multirow{11}{*}{\rotatebox[origin=c]{90}{WSVM}}} 
	& Advertisement	     & 0.94	 & 0.96  & 0.80  & \textbf{0.88} & 7	& 0.92  & 0.99  & 0.45  & 0.67            \\
	& Buzz	             & 0.94	 & 0.96  & 0.87  & \textbf{0.91} & 14	& 0.96  & 0.99  & 0.81  & 0.89            \\
	& Clean (Musk)	     & 1.00  & 1.00  & 0.99  & \textbf{0.99} & 5	& 1.00  & 1.00  & 0.98  & 0.99            \\
	& Cod-rna		     & 0.94  & 0.97  & 0.95  & \textbf{0.96} & 9	& 0.96  & 0.96  & 0.96  & \textbf{0.96}   \\
	& EEG Eye State	     & 0.87  & 0.89  & 0.86  & \textbf{0.88} & 6    & 0.88  & 0.90  & 0.86  & \textbf{0.88}   \\
	& Forest (Class 5)   & 0.92  & 0.92  & 0.90  & 0.91          & 33   & 1.00  & 1.00  & 0.86  & \textbf{0.93}   \\
	& Hypothyroid		 & 0.98  & 0.98  & 0.75  & \textbf{0.86} & 4	& 0.99  & 1.00  & 0.75  & \textbf{0.86}   \\
	& ISOLET	         & 0.99  & 1.00  & 0.85  & \textbf{0.92} & 11   & 0.99  & 1.00  & 0.85  & \textbf{0.92}   \\
	& Letter	         & 0.99  & 0.99  & 0.96  & \textbf{0.99} & 8	& 1.00  & 1.00  & 0.97  & \textbf{0.99}   \\
	& Nursery		     & 1.00  & 0.99  & 0.98  & 0.99          & 10   & 1.00  & 1.00  & 1.00  & \textbf{1.00}   \\
	& Protein homology		     & 1.00	 & 1.00	 & 0.87  & \textbf{0.92} & 18   & 1.00  & 1.00  & 0.80  & 0.89            \\
	& Ringnorm	         & 0.98	 & 0.97  & 0.99  & \textbf{0.98} & 6	& 0.98  & 0.99  & 0.98  & \textbf{0.98}   \\
	& Twonorm	         & 0.97  & 0.98  & 0.97  & 0.97          & 6	& 0.98  & 0.98  & 0.99  & \textbf{0.98}   \\
\hline
\end{tabular}}
\end{table}


\subsection{\mlsvm-AMG sparsity preserving coarsening}\label{sec:r1amg}
We have experimented with the light version of \mlsvm-AMG in which instead of computing a linear combination of $f$-level points to get $c$-level points (see Eq. \ref{eq:pointagg}), we prolongate the seed to be a corresponding coarse point in attempt to preserve the sparsity of data points. 
In terms of quality of classifiers, the performance measures of this method are similar to that of \mlsvm-IIS and in most cases (see Tables \ref{table5}-\ref{table6}) 
are faster. However, for Buzz and Cod-rna datasets, although \mlsvm-AMG performs faster, it results in a lower sensitivity and specificity (see Table \ref{table6}) for SVM, and higher sensitivity and specificity for WSVM (see Table \ref{table6}) compared to \mlsvm-IIS. For Protein dataset, the sensitivity and specificity are improved compared to \mlsvm-IIS (see Table \ref{table6}). 

\begin{table}[H]
\centering
\caption{Comparison of computational time for single- (LibSVM) and multilevel (\mlsvm-IIS and sparse \mlsvm-AMG) solvers in seconds. Presented values include running time in seconds for both WSVM and SVM with model selection.}
\label{table5}
  \scalebox{.8}{
    \begin{tabular}{lccc}
    \hline
Dataset	&	 \mlsvm-IIS & Sparse \mlsvm-AMG & Single-level  \\ \hline
	Advertisement	   & 196	& {\bf 91} & 412	  \\
	Buzz               & 2329 & {\bf 957}	& 70452	  \\
	Clean (Musk)	   & 30 & {\bf 6}	& 167	  \\
	Cod-rna	           & 172 & \bf{92}	& 1611	  \\
	EEG Eye State	   & 51  & \bf{45}	& 447	  \\
	Forest (Class 5)   & 13785 & \bf{13328} & 352500  \\
	Hypothyroid	       & \bf{3}	& \bf{3}	    & 5	      \\
	ISOLET	           & 69 & \bf{64}	& 1367	  \\
	Letter	           & 45 & \bf{18}	& 333	  \\
	Nursery	           & 63 & \bf{33}	& 519	  \\
	Protein homology           & \bf{1564} & 1597  & 73311   \\
	Ringnorm	       & \bf{4}	& 5     & 42	  \\
	Twonorm	           & \bf{4}	& {\bf 4}    & 45	  \\
\hline
\end{tabular}
}
\end{table}

\begin{table}[H]
\centering
\caption{Performance measures of regular and weighted \mlsvm-AMG.  Column 'Depth' shows the number of levels in the multilevel hierarchy which is independent of SVM type.}
\label{table6}
\scalebox{.8}{
    \begin{tabular}{lcccc|ccccc}                  
        \hline
    & \multicolumn{4}{c}{regular \mlsvm-AMG}         & \multicolumn{4}{|c}{weighted \mlsvm-AMG}  & \multicolumn{1}{c}{Depth} \\ \cline{2-5}   \cline{6-9} 
			   Dataset  		& ACC	& SN	& SP	& G-mean   	  & ACC		& SN	& SP	& G-mean & 	\\\hline
	Advertisement	    & 0.95  & 0.99  & 0.64  & \bf{0.86}        & 0.95  & 0.99  & 0.64  & \bf{0.86}   & 2     \\
	Buzz                & 0.87  & 0.89  & 0.79  & 0.83             & 0.93  & 0.95  & 0.85  & \bf{0.90}   & 8     \\
	Clean               & 0.99  & 1.00  & 0.98  & \bf{0.99}        & 0.99  & 1.00  & 0.98  & \bf{0.99}   & 4     \\
	Cod-rna	            & 0.86  & 0.85  & 0.88  & 0.87             & 0.89  & 0.89  & 0.90  & \bf{0.90}   & 6     \\
	EEG Eye State       & 0.87  & 0.88  & 0.85  & \bf{0.86}        & 0.87  & 0.88  & 0.85  & \bf{0.86}   & 4     \\
	Forest (Class 5)    & 0.97  & 0.98  & 0.79  & 0.88             & 0.96  & 0.97  & 0.82  & \bf{0.89}   & 9     \\
	ISOLET	            & 0.99  & 1.00  & 0.83  & \bf{0.91}        & 0.99  & 1.00  & 0.83  & \bf{0.91}   & 3     \\
	Letter	            & 0.99  & 0.99  & 0.95  & \bf{0.97}        & 0.99  & 0.99  & 0.93  & 0.96        & 5     \\
	Nursery	            & 0.99  & 0.99  & 1.00  & 0.99             & 1.00  & 1.00  & 1.00  & \bf{1.00}   & 4     \\
	Protein	homology            & 0.97  & 0.97  & 0.86  & \bf{0.91}        & 0.97  & 0.97  & 0.85  & \bf{0.91}   & 5     \\
	Ringnorm            & 0.98  & 0.98  & 0.98  & \bf{0.98}        & 0.98  & 0.99  & 0.98  & \bf{0.98}   & 3     \\
	Twonorm	            & 0.98  & 0.97  & 0.98  & \bf{0.98}        & 0.98  & 0.97  & 0.98  & \bf{0.98}   & 3     \\ 
\hline
\end{tabular}}
\end{table}

%

We perform the sensitivity analysis of the order of interpolation denoted by $r$ (see Eq. \ref{eq:interp}), the maximum number of fractions a point in $F$ can be divided into, and compare the performance measures and computational time in Table \ref{tbl:interpolation_order_effect}. As $r$ increases, the performance measures such as G-mean are improving until they do not stop  changing for larger $r$. For example, for Buzz dataset, the G-mean is not changing for larger $r=6$. The presented results are computed without advancements FF and FS (see Section 3.3). Using these techniques, we obtain G-mean 0.95 with $r=1$ for Buzz data set. Higher interpolation orders increase the time but produce the same quality on that data set. 

\begin{table}[h]
\centering
\caption{Sensitivity analysis of interpolation order $r$ in \mlsvm-AMG for Buzz data set.}
\label{tbl:interpolation_order_effect}
\scalebox{.8}{
    \begin{tabular}{lcccccccc}
        \hline
         & Metric	&	$r=1$	&	$r=2$	&	$r=3$	& $r=4$ & $r=5$ &	$r=6$	&	$r=10$	\\	\hline
          \multirow{4}[0]{*}{{ {\mlsvm-AMG SVM}}}
        & G-mean  & 0.26  & 0.33  & 0.56  & 0.83  & 0.90         & \bf{0.91}  & 0.89 \\
        & SN      & 0.14  & 0.33  & 0.68  & 0.89  & \bf{0.98}    & 0.97       & 0.95 \\
        & SP      & 0.47  & 0.34  & 0.47  & 0.79  & 0.82         & \bf{0.86}  & 0.82 \\
        & ACC     & 0.21  & 0.33  & 0.64  & 0.87  & \bf{0.95}    & \bf{0.95}  & 0.94 \\
        \hline
          \multirow{4}[0]{*}{{ {\mlsvm-AMG WSVM}}} 			
        & G-mean  & 0.26  & 0.40  & 0.60  & 0.90  & 0.93        & 0.93       & \bf{0.94} \\ 
        & SN      & 0.14  & 0.32  & 0.74  & 0.95  & \bf{0.98}   & \bf{0.98}  & \bf{0.98} \\
        & SP      & 0.47  & 0.5   & 0.48  & 0.85  & 0.88        & \bf{0.89}  & \bf{0.89} \\ 
        & ACC     & 0.21  & 0.35  & 0.69  & 0.93  & 0.96        & \bf{0.97}  & \bf{0.97} \\ 
            \hline
        & time(sec.)	&	389	&	541	&	659	& 957 & 1047   &	1116	&	1375	\\	\hline
        \end{tabular}
}
\end{table}

\begin{figure*}[H]
 \centering
  \begin{tabular}{cc}
           \includegraphics[scale=0.45]{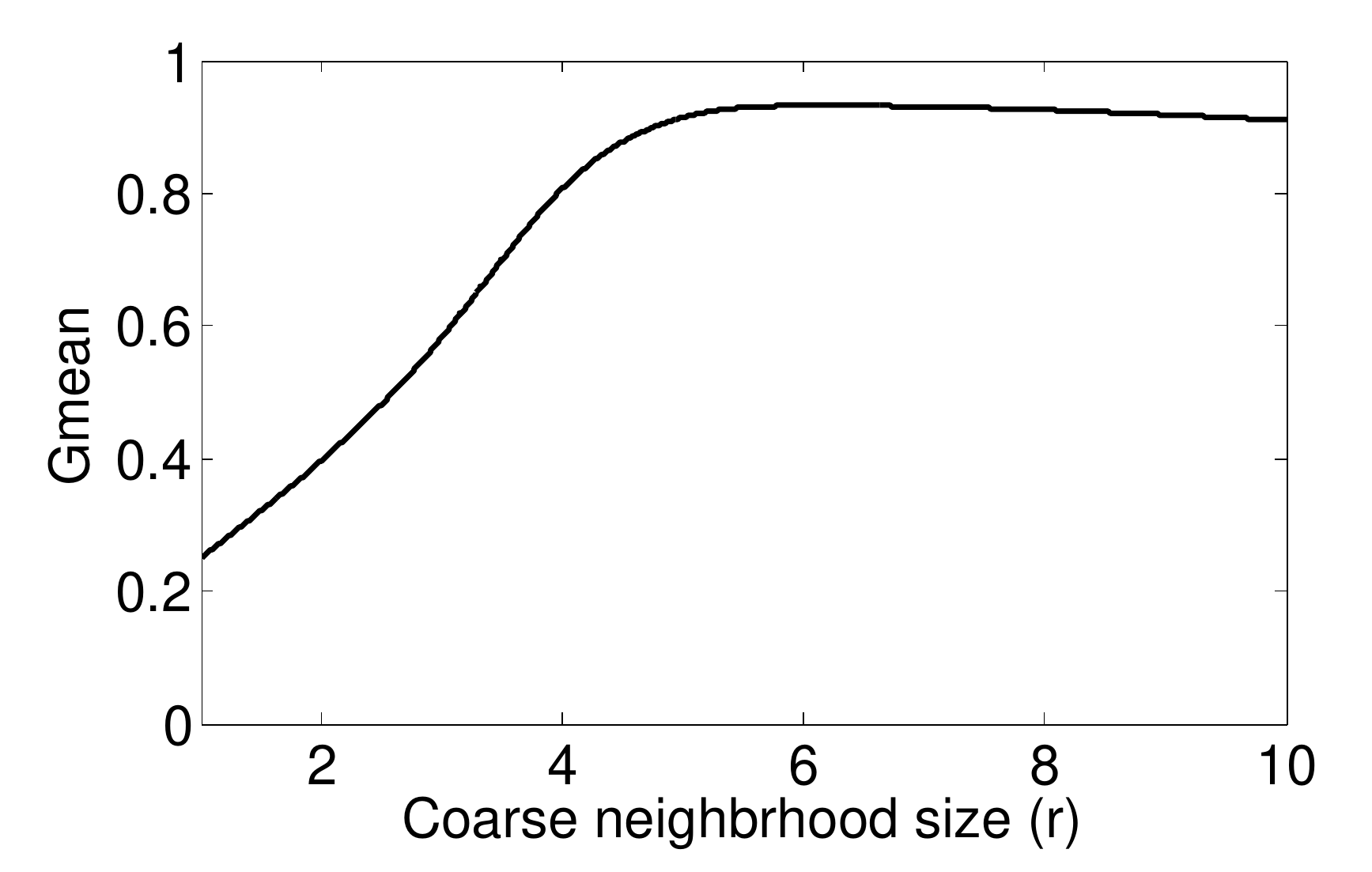} &\includegraphics[scale=0.47]{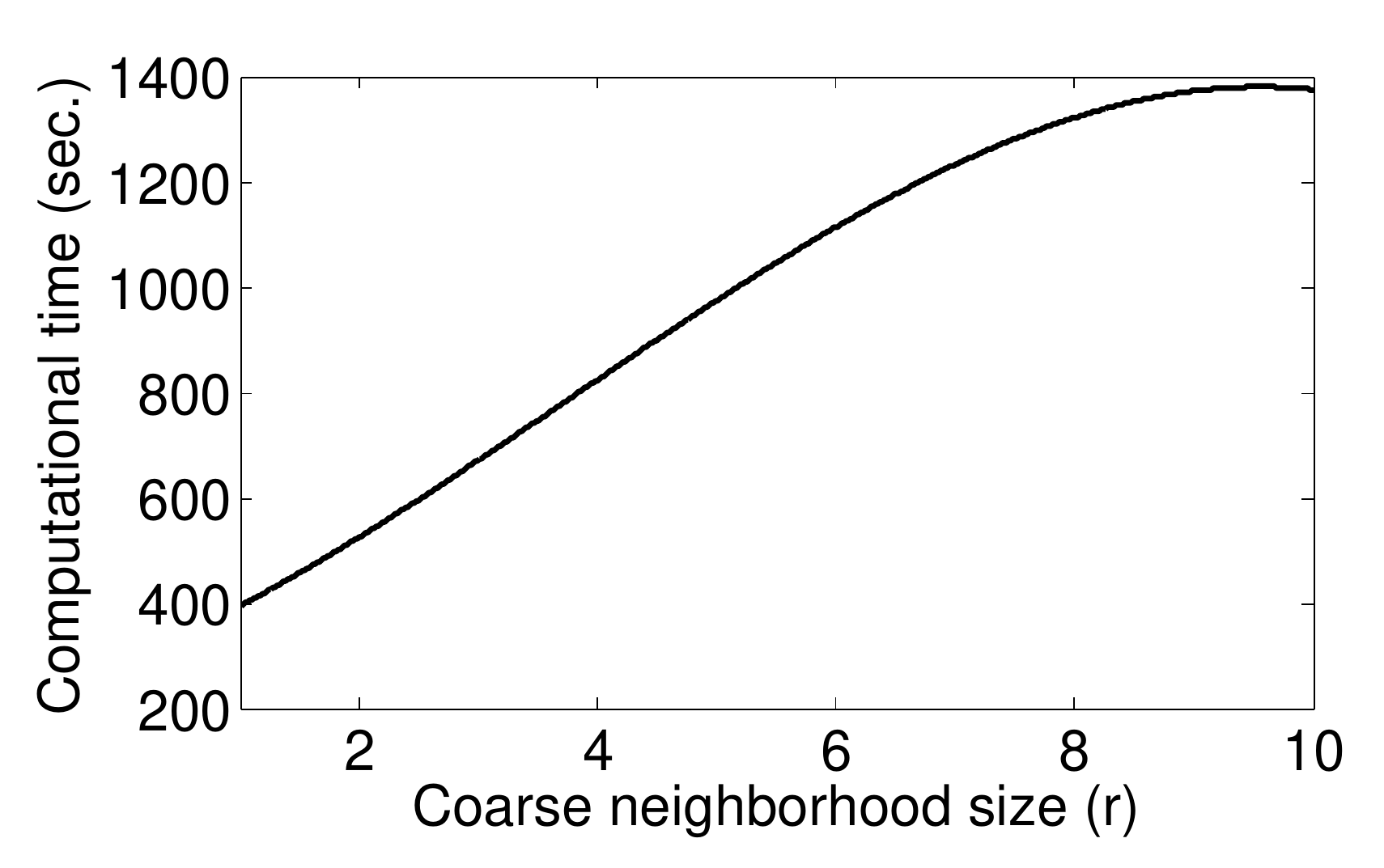} \\	
           (a)  Letter26 &  (b) Hypothyroid
\end{tabular}
\caption{Sensitivity of AMG-(W)SVM in terms of G-mean metric and computational time to $r$ for Buzz data}
\label{fig_Sen1}

\end{figure*}

\subsection{Full \mlsvm-AMG coarsening}
The best version of full \mlsvm-AMG coarsening whose results are reported, chooses the best model from different scales (see Sec. \ref{sec:mod-scales}). 
For this type of \mlsvm-AMG, all experiments on all data sets have been executed on a single machine with CPU Intel Xeon E5-2665 2.4 GHz and 64 GB RAM. The framework runs in sequential mode. The FLANN library is used to generate the approximated $k$-NN graph for $k=10$. Once it is generated for the whole data set, its result is saved and reused. In all experiments all data points are randomly reordered as well as for each $k$-fold, the indices from the original test data are removed and reordered, so \emph{no order in which points are entered into QP solver affects the solution}. 
Each experiment includes a full $k$-fold cross validation. 
The average performance measures over 5 experiments each of which includes 10-fold cross validation are demonstrated in Tables \ref{tab:new_results_agg_seeds_performance_measures} and \ref{tab:new_results_agg_seeds_time}.
%
The best model among all the levels for each fold of cross validation is selected using \emph{validation} data (see Sec. \ref{ref_model_selection}).
Using the best model, the performance measures over the test data are calculated and reported as the final performance for this specific fold of cross validation.

For the purpose of comparison, the results of previous work using validation techniques CS, CCkF without partitioning the training data during the refinement \cite{sadrfaridpour2016algebraic} are presented in Table \ref{tab:alleval_ESSAN}. 
\begin{table*}[htb]
\setlength{\tabcolsep}{3.8pt}
  \caption{Performance measures and running time (in seconds) for weighted single level SVM (LibSVM), and weighted mlsvm-AMG on benchmark data sets in \cite{blake1998uci} \emph{without partitioning}.}
  \label{tab:alleval_ESSAN}
  \begin{center}
 \scalebox{.8}{
    \begin{tabular}{lccccc|ccccc}
      \hline
      &
      \multicolumn{5}{c|}{Single level WSVM} &
      \multicolumn{5}{c}{mlsvm-AMG} \\ \cline{2-11}
        Dataset           & ACC      & SN   & SP   & G-mean & Time & ACC      & SN   & SP   & G-mean   & 	Time	 \\
      \hline
        Advertisement   & 0.92  & 0.99  & 0.45  & 0.67       & 231     & 0.83 & 0.92 & 0.81 & \bf{0.86}  & \bf{213}   \\
        Buzz            & 0.96  & 0.99  & 0.81  & 0.89       & 26026   & 0.88 & 0.97 & 0.86 & \bf{0.91}  & \bf{233}   \\
        Clean (Musk)    & 1.00  & 1.00  & 0.98  & \bf{0.99}  & 82      & 0.97 & 0.97 & 0.97 & 0.97       & \bf{7}     \\
        Cod-RNA         & 0.96  & 0.96  & 0.96  & \bf{0.96}  & 1857    & 0.94 & 0.97 & 0.92 & 0.95       & \bf{102}   \\
        Forest          & 1.00  & 1.00  & 0.86  & \bf{0.92}  & 353210  & 0.88 & 0.92 & 0.88 & 0.90       & \bf{479}   \\
        Hypothyroid     & 0.99  & 1.00  & 0.75  & 0.86       & 3       & 0.98 & 0.83 & 0.99 & \bf{0.91}  & \bf{3 }    \\
        ISOLET		  & 0.99  & 1.00  & 0.85  & 0.92       & 1367    & 0.99 & 0.89 & 1.00 & \bf{0.94}  & \bf{66}	\\      
        Letter          & 1.00  & 1.00  & 0.97  & \bf{0.99}  & 139     & 0.98 & 1.00 & 0.97 & \bf{0.99}  & \bf{12 }   \\
        Nursery         & 1.00  & 1.00  & 1.00  & \bf{1.00}  & 192     & 1.00 & 1.00 & 1.00 & \bf{1.00}  & \bf{2  }   \\
        Ringnorm        & 0.98  & 0.99  & 0.98  & \bf{0.98}  & 26      & 0.98 & 0.98 & 0.98 & \bf{0.98}  & \bf{2 }    \\
        Twonorm         & 0.98  & 0.98  & 0.99  & \bf{0.98}  & 28      & 0.98 & 0.98 & 0.97 & \bf{0.98}  & \bf{1  }   \\

      \hline
    \end{tabular}
  }
  \end{center}
\normalsize
\end{table*}

The results using validation techniques FF, FS with partitioning the training data during the refinement phase are presented in Tables \ref{tab:new_results_agg_seeds_performance_measures}, \ref{tab:new_results_agg_seeds_time}. We compare our performance and quality with those obtained by LibSVM, DC-SVM, and Ensemble SVM. All results are related to WSVM. The ``Single level WSVM'' column in Table \ref{tab:new_results_agg_seeds_performance_measures} represents the weighted SVM results produced by LibSVM. The LibSVM solver is slow but it produces almost the best G-mean results over our experimental datasets except Advertisement, Buzz, and Forest.
The DC-SVM \cite{pmlr-v32-hsieha14} produces better G-mean on 4 datasets compare to LibSVM (see Table \ref{tab:new_results_agg_seeds_performance_measures}) but has lower G-mean on 4 other datasets. We choose DC-SVM
not only because it has a hierarchical framework (with different principles of (un)coarsening) but also because it significantly outperforms other hierarchical techniques which are typically fast but not of high quality. 

The mlsvm-AMG demonstrates significantly better computation time than DC-SVM on almost all datasets (see Table \ref{tab:new_results_agg_seeds_performance_measures}). 
Furthermore, mlsvm-AMG classification quality is significantly better on both Advertisement and Buzz datasets compared to LibSVM.
In addition, the comparison between DC-SVM and mlsvm-AMG shows that the latter has higher G-mean for Advertisement, Buzz, Clean, Cod, Ringnorm, and Twonorm datasets.
A better performance of DC-SVM is observed on Forest dataset if \mlsvm-AMG is applying partitioning, i.e., when the number of support vectors is big. However, in another version of multilevel framework with validation techniques CS, CCkF without partitioning the training data during the refinement, the G-mean raises to 0.90 (see Table \ref{tab:alleval_ESSAN}). 
It is interesting to note that the dimensionality of Advertisement dataset is the main source of complexity for the parameter fitting in both LibSVM and \mlsvm-AMG.
All versions of multilevel SVMs produce G-mean 0.90 for this dataset which is significantly higher than that of LibSVM which is 0.67.
The results for this dataset are not significantly different for DC-SVM which is, however, 3 times slower than full \mlsvm-AMG and 6 times slower than sparse \mlsvm-AMG.
\begin{table*}[!h]
  \setlength{\tabcolsep}{4pt}
  \caption{Performance measures for single level WSVM (LibSVM), DC-SVM and \mlsvm-AMG on benchmark data sets using partitioning and FF, FS validation techniques.} 
  \label{tab:new_results_agg_seeds_performance_measures}
  \begin{center}
    \scalebox{.8}{
    \begin{tabular}{lcccc|cccc|cccc}
      \hline
      %
       &
      \multicolumn{4}{c|}{Single level WSVM} &
      \multicolumn{4}{c|}{DC-SVM} &
      \multicolumn{4}{c}{mlsvm-AMG} \\ \cline{2-13}
      Datasets           & ACC      & SN   & SP   & G-mean & ACC      & SN   & SP   & G-mean & ACC      & SN   & SP   & G-mean\\
      \hline
	Advertisement   & 0.92 & 0.99 & 0.45 & 0.67         & 0.95 & 0.83 & 0.97 & 0.90   & 0.95 & 0.85 & 0.96 & \bf{0.91}         \\
	Buzz            & 0.96 & 0.99 & 0.81 & 0.89         & 0.96 & 0.88 & 0.97 & 0.92        & 0.94 & 0.95 & 0.94 & \bf{0.95}    \\
	Clean (Musk)    & 1.00 & 1.00 & 0.98 & \bf{0.99}    & 0.96 & 0.91 & 0.97 & 0.94        & 0.99 & 0.99 & 0.99 & \bf{0.99}    \\
	Cod-RNA         & 0.96 & 0.96 & 0.96 & \bf{0.96}    & 0.93 & 0.93 & 0.94 & 0.93        & 0.93 & 0.97 & 0.91 & \bf{0.94}    \\
	Forest          & 1.00 & 1.00 & 0.86 & 0.92         & 1.00 & 0.88 & 1.00 & \bf{0.94}   & 0.77 & 0.96 & 0.80 & 0.88         \\
	Letter          & 1.00 & 1.00 & 0.97 & 0.99         & 1.00 & 1.00 & 1.00 & \bf{1.00}   & 0.98 & 0.99 & 0.98 & 0.99         \\
	Nursery         & 1.00 & 1.00 & 1.00 & \bf{1.00}    & 1.00 & 1.00 & 1.00 & \bf{1.00}   & 1.00 & 1.00 & 1.00 & \bf{1.00}    \\
	Ringnorm        & 0.98 & 0.99 & 0.98 & \bf{0.98}    & 0.95 & 0.92 & 0.98 & 0.95        & 0.98 & 0.98 & 0.98 & \bf{0.98}    \\
	Twonorm         & 0.98 & 0.98 & 0.99 & \bf{0.98}    & 0.97 & 0.98 & 0.96 & 0.97        & 0.98 & 0.98 & 0.97 & \bf{0.98}    \\
      \hline
    \end{tabular}}
  \end{center}
  \normalsize
\end{table*}

The computational time in seconds is demonstrated in Table \ref{tab:new_results_agg_seeds_time}. Our experiments exhibit significant performance improvement.
\begin{table*}[!h]
  \normalsize
  \caption{Computational time in seconds for single level WSVM (LibSVM), DC-SVM and mlsvm-AMG.} 
  \label{tab:new_results_agg_seeds_time}
  \begin{center}
    \scalebox{.8}{
      \begin{tabular}{lccc}
	\hline
	Dataset           & Single level WSVM & DC-SVM   & mlsvm-AMG\\
	\hline
	    Advertisement   & 231        & 610      &  \bf{213}     \\
	    Buzz            & 26026      & 2524     &  \bf{31 }     \\
	    Clean (Musk)    & \bf{82}    & 95       &  94           \\
	    Cod-RNA         & 1857       & 420      &  \bf{13  }    \\
	    Forest          & 353210     & 19970    &  \bf{948}    \\
	    Letter          & 139        & 38       &  \bf{30}      \\
	    Nursery         & 192        & 49       &  \bf{2 }      \\
	    Ringnorm        & 26         & 38       &  \bf{2 }      \\
	    Twonorm         & 28         & 30       &  \bf{1 }      \\
	\hline
      \end{tabular}}
  \end{center}
  \normalsize
\end{table*}
%
%
%
\subsubsection{Large datasets}

Large datasets SUSY and Higgs are available at UCI repository \cite{Lichman:2013}. The MNIST8M was downloaded from LibSVM data repository. Half of each class was randomly sampled to make classification more difficult. All the methods (our and competitors') are benchmarked using Intel Xeon (E5-2680v3) with 128Gb memory. 

The experiments with DC-SVM have not been finished after 3 full days of running and its performance is not presented because of unrealistic slowness of the method. Therefore, it is not comparable with \mlsvm-AMG on large datasets. The LibSVM performs slower than DC-SVM on these datasets and is also not presented. Although, fast linear SVM solvers are beyond the scope of this work, we compare the \mlsvm-AMG with the LibLinear  \cite{fan2008liblinear} that is significantly faster than both DC-SVM and LibSVM. We note that linear SVM solvers can also be used as the refinement in multilevel frameworks. However, in practice, we do not observe a need for this because nonlinear SVM refinement is already fast enough in our multilevel framework. 

\begin{figure}
    \centering
    \includegraphics[width=1\textwidth]{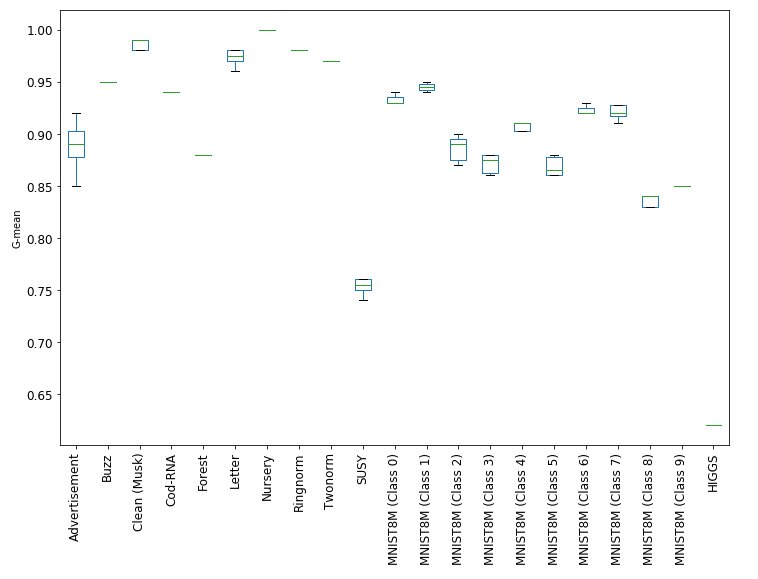}
    \caption{Each boxplot (horizontal axis) shows variability of the G-mean (vertical axis). A small standard deviation is observed in all cases.}
    \label{fig:stdev}
\end{figure}

The results for performance measures and computational time are presented in Tables \ref{tab:large_datasets_mlsvm_AMG_performance_measures}, and \ref{tab:time_larger_datasets}.
The \mlsvm-AMG produces higher G-means on SUSY, HIGGS, and 8 (out of 10) of classes in the MINST8M datasets. 
On classes 8, 5, and 9 of MNIST8M we have an improvement of 24\%, 6\% and 5\%, respectively.
On the average, the G-mean for all larger datasets are 5\% higher for \mlsvm-AMG in comparison to LibLinear. 
The \mlsvm-AMG is faster than LibLinear on SUSY and HIGGS datasets and slower on MNIST8M dataset. However, this slowness is eliminated if linear SVM solver is used in the refinement. The results for seven classes of Forest dataset are presented in Table \ref{tbl:forest_classes_results}. The statistics of G-mean variability is presented in Figure \ref{fig:stdev} which confirms the robustness of the proposed method. 
\begin{figure}
    \centering
    \includegraphics[width=0.85\textwidth]{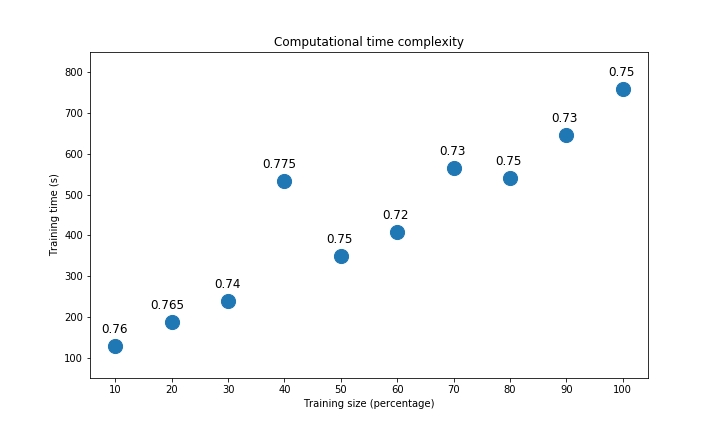}
    \caption{Scalability of \mlsvm-AMG on growing training set of SUSY dataset. Each point represents the training time (vertical axis) when a certain part of the full training set (horizontal axis) is used. The numbers above points represent the G-mean performance measure. For example, if we use 60\% of the training set to train the model, the running time is about 400 seconds, and the G-mean is 0.72.}
    \label{fig:scal}
\end{figure}
\begin{figure}
    \centering
    \includegraphics[width=0.85\textwidth]{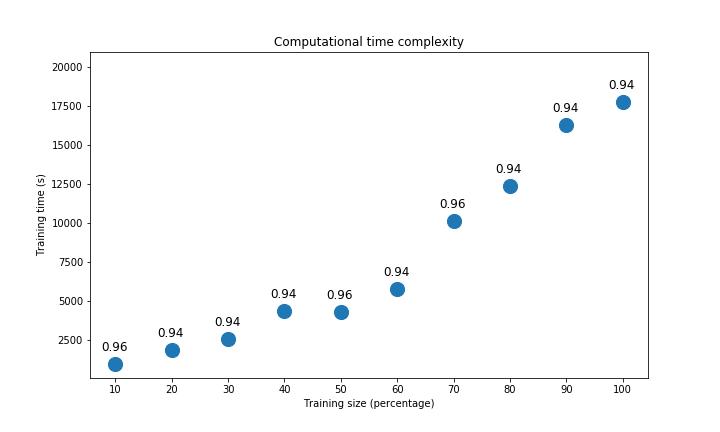}
    \caption{Scalability of \mlsvm-AMG on growing training set of MNIST8M dataset using class 1. The 5M data points from the MNIST8M dataset are sampled to create a similar size comparison with SUSY dataset for a larger number of features.} 
    \label{fig:scal_mnist_c1}
\end{figure}

\begin{figure}
    \centering
    \includegraphics[width=1\textwidth]{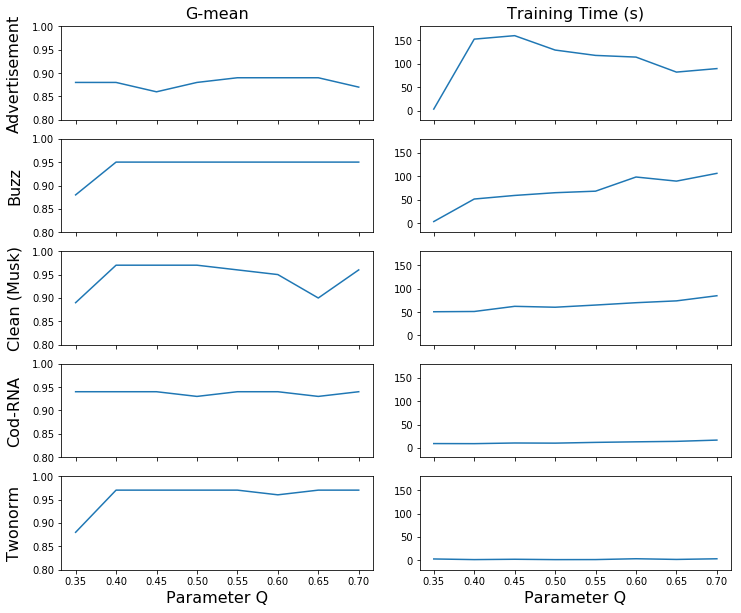}
    \caption{The \mlsvm-AMG using parameter $Q\in [0.35,...,0.7]$ generates the best results on the benchmark data sets.}
    \label{fig:scal_param_Q}
\end{figure}

In many cases, we observe a faster than linear behavior of our framework. An example is shown in Figure \ref{fig:scal}. When we use only a part of the dataset SUSY for training the model (horizontal axis), the computational time (vertical axis) is increasing slower than linearly. Such behavior can be observed when the number of support vectors is relatively small which is one of the main assumptions of this method. Another example with a larger number of features for MNIST8M is presented in Figure \ref{fig:scal_mnist_c1}.

The robustness of parameter $Q$ (see Alg. \ref{alg:coarsening}, line \ref{ln:paramQ}), which determines the size of the coarse level is also an important question. In AMG and AMG-inspired algorithms, a typical setting is to make $Q\in [0.4, \dots, 0.6]$ unless a special reason for a faster aggregation allows more aggressive compression of the problem without significant loss in the solution quality. Here we observe that a similar range for $Q$ is generally robust (see Fig. \ref{fig:scal_param_Q}). In general, in multilevel learning, over-compression with too small $Q$ is not recommended unless we know that a data is easily separable (or well clustered). 

Finally, we present the computational time in terms of the amount of work per unit for all datasets in Table \ref{tbl:complex}. In ``$ \frac{\mu s}{point}$'' and ``$ \frac{\mu s}{value}$'' columns, we present the computational time in microseconds per data point and one feature value in data point, respectively.
\begin{table}[htb]
\centering
\caption{Larger benchmark data sets.}
\label{tbl_larger_dataset_info}
\scalebox{.8}{
    \begin{tabular}{lccccccc}         
	\hline
		Dataset	          &	$\epsilon$    & $n_f$        & $|\JJ|$	   & $|\CC^+|$	    & $|\CC^-|$	\\ \hline
		SUSY              & 0.54         & 18           & 5000000      & 2287827      & 2712173    \\
        MNIST8M (Class 0) & 0.90         & 784          & 4050003      & 399803       & 3650200    \\
		MNIST8M (Class 1) & 0.89         & 784          & 4050003      & 455085       & 3594918    \\
		MNIST8M (Class 2) & 0.90         & 784          & 4050003      & 402165       & 3647838    \\
		MNIST8M (Class 3) & 0.90         & 784          & 4050003      & 413843       & 3636160    \\
		MNIST8M (Class 4) & 0.90         & 784          & 4050003      & 394335       & 3655668    \\
		MNIST8M (Class 5) & 0.91         & 784          & 4050003      & 365918       & 3684085    \\
		MNIST8M (Class 7) & 0.90         & 784          & 4050003      & 399465       & 3650538    \\
		MNIST8M (Class 6) & 0.90         & 784          & 4050003      & 422888       & 3627115    \\
		MNIST8M (Class 8) & 0.90         & 784          & 4050003      & 394943       & 3655060    \\
		MNIST8M (Class 9) & 0.90         & 784          & 4050003      & 401558       & 3648445    \\
		HIGGS             & 0.53         & 28           & 11000000     & 5170877      & 5829123    \\	\hline
	\end{tabular}
	}
\end{table}
%
%
%
%
\begin{table*}[!h]
  \setlength{\tabcolsep}{3.8pt}
  \caption{Performance measures for single level WSVM (LibLinear), DC-SVM/LibSVM and mlsvm-AMG on larger benchmark data sets using partitioning and FF, FS validation techniques.} 
  \label{tab:large_datasets_mlsvm_AMG_performance_measures}
  \begin{center}
    \scalebox{.8}{
    \begin{tabular}{lcccc|cccc|cccc}
      \hline
      &
      \multicolumn{4}{c|}{LibLinear} &
      \multicolumn{4}{c|}{DC-SVM and LibSVM} &
      \multicolumn{4}{c}{mlsvm-AMG} \\ \cline{2-13} 
		Dataset           & ACC      & SN   & SP   & G-mean &\multicolumn{4}{c|}{} & ACC      & SN   & SP   & G-mean\\
      \hline
        SUSY               & 0.69   & 0.61   & 0.76   & 0.68          &\multicolumn{4}{c|}{}              & 0.75   & 0.71   & 0.78   & \bf{0.74}    \\
        MNIST8M (Class 0)  & 0.98   & 0.90   & 0.99   & \bf{0.95}     &\multicolumn{4}{c|}{}              & 0.94   & 0.93   & 0.94   & \bf{0.95}  \\
        MNIST8M (Class 1)  & 0.98   & 0.93   & 0.99   & \bf{0.96}     &\multicolumn{4}{c|}{}              & 0.94   & 0.95   & 0.94   & 0.95  \\
        MNIST8M (Class 2)  & 0.97   & 0.77   & 0.99   & 0.87          &\multicolumn{4}{c|}{}              & 0.91   & 0.87   & 0.91   & \bf{0.89}  \\
        MNIST8M (Class 3)  & 0.96   & 0.70   & 0.98   & 0.83          &\multicolumn{4}{c|}{}              & 0.88   & 0.86   & 0.89   & \bf{0.88}  \\
        MNIST8M (Class 4)  & 0.97   & 0.81   & 0.99   & 0.90     &\multicolumn{4}{c|}{Stopped or failed after 3 days}       & 0.91   & 0.92   & 0.91   & \bf{0.91}  \\
        MNIST8M (Class 5)  & 0.96   & 0.64   & 0.99   & 0.80          &\multicolumn{4}{c|}{without any result}              & 0.84   & 0.91   & 0.83   & \bf{0.86}  \\
        MNIST8M (Class 6)  & 0.98   & 0.86   & 0.99   & 0.92          &\multicolumn{4}{c|}{}              & 0.94   & 0.90   & 0.94   & \bf{0.93}  \\
        MNIST8M (Class 7)  & 0.98   & 0.85   & 0.99   & 0.91          &\multicolumn{4}{c|}{}              & 0.93   & 0.90   & 0.93   & \bf{0.92}  \\
        MNIST8M (Class 8)  & 0.92   & 0.36   & 0.98   & 0.60          &\multicolumn{4}{c|}{}              & 0.82   & 0.87   & 0.81   & \bf{0.84}  \\
        MNIST8M (Class 9)  & 0.94   & 0.64   & 0.97   & 0.80          &\multicolumn{4}{c|}{}              & 0.81   & 0.91   & 0.79   & \bf{0.85}  \\
        HIGGS              & 0.54   & 0.55   & 0.54   & 0.54          &\multicolumn{4}{c|}{}              & 0.62   & 0.61   & 0.63   & \bf{0.62}  \\
      \hline
    \end{tabular}}
  \end{center}
\end{table*}

\begin{table}[!h]
  \normalsize
  \setlength{\tabcolsep}{2pt}
  \caption{Computational time in seconds for single level WSVM (LibLinear), DC-SVM/LibSVM and mlsvm-AMG on larger benchmark data sets} 
  \label{tab:time_larger_datasets}
  \begin{center}
    \scalebox{.8}{
      \begin{tabular}{lc|c|c}
    \hline
    Dataset           & LibLinear & DC-SVM and LibSVM   & mlsvm-AMG\\
    \hline
        SUSY                      & 1300        &                         &  \bf{1116}     \\
        MNIST8M (Class 0)         & \bf{1876}   &                         & 11411      \\
        MNIST8M (Class 1)         & \bf{859}    &                         & 15441     \\
        MNIST8M (Class 2)         & \bf{1840}   &                         & 17398      \\
        MNIST8M (Class 3)         & \bf{2362}   &                         & 10547      \\
        MNIST8M (Class 4)         & \bf{1448}   & Stopped or failed after 3 days    & 13014      \\
        MNIST8M (Class 5)         & \bf{2360}   & without any result      & 13353      \\
        MNIST8M (Class 6)         & \bf{1628}   &                         & 10092      \\
        MNIST8M (Class 7)         & \bf{1747}   &                         & 16789      \\
        MNIST8M (Class 8)         & \bf{2626}   &                         & 17581      \\
        MNIST8M (Class 9)         & \bf{1650}   &                         & 21611      \\
        HIGGS                     &  4406       &                         & \bf{3283}     \\
    \hline                                                    
      \end{tabular}}
  \end{center}
  \normalsize
\end{table}

\begin{table}[!h]
\normalsize
\centering
\caption{Performance measures and running time (in seconds) for all classes of Forest dataset using full mlsvm-AMG. }
\label{tbl:forest_classes_results}
\scalebox{.8}{
    \begin{tabular}{lcccccc}
    \hline
        Dataset		&	ACC	 	 &	SN		&	SP		& G-mean &	PPV	 	&	 	Time	\\ \hline
        Class 1     &	0.73	 &	0.79	&	0.69	&	0.74   &	0.60	&		926	    \\ 
        Class 2     &	0.70	 &	0.78	&	0.62	&	0.70   &	0.67	&		215	    \\ 
        Class 3     &	0.90	 &	0.99	&	0.90	&	0.94   &	0.39	&		1496    \\ 
        Class 4    	&	0.92	 &	1.00	&	0.92	&	0.96   &	0.99	&		3231	\\ 
        Class 5		&	0.80	 &	0.96	&	0.80	&	0.88   &	0.07	&		948	\\
        Class 6     &	0.86	 &	0.95	&	0.95	&	0.90   &	0.17	&		2972	\\ 
        Class 7     &	0.91	 &	0.87	&	0.91	&	0.89   &	0.28	&		2269	\\ 
    \hline
    \end{tabular}
}
\end{table}
\begin{table}[!h]
\centering
\caption{Complexity Analysis}
\label{tbl:complex}
\scalebox{.8}{
\begin{tabular}{lrrrrrrrc}
\hline
Dataset                      & $|\JJ|$    & $n_f$   & $|\JJ|\cdot n_f$             
& \resizebox{0.08\hsize}{!}{$ \frac{\mu s}{point} $} 
& \resizebox{0.08\hsize}{!}{$ \frac{\mu s}{value}$} \\ 
\hline
Nursery                      & 13K     & 19      & 246.2K       & 232       & 12     \\ 
Twonorm                      & 7.4K    & 20      & 148K         & 405       & 20     \\ 
Ringnorm                     & 7.4K    & 20      & 148K         & 541       & 27     \\ 
Letter                       & 20K     & 16      & 320K         & 100       & 6     \\ 
Cod-rna                      & 59.5K   & 8       & 476.3K       & 100       & 13     \\ 
Clean (Musk)                 & 6.6K    & 166     & 1.1M         & 909       & 6      \\
Advertisement                & 3.3K    & 1558    & 5.1M         & 31107     & 20       \\
Buzz                         & 140.7K  & 77      & 10.8M        & 1628      & 21      \\
Forest                       & 581K    & 54      & 31.4M        & 207       & 4     \\
Susy                         & 5M      & 18      & 90M          & 223       & 12     \\
Higgs                        & 11M     & 28      & 308M         & 298       & 11     \\
mnist 4M                     & 4.1M    & 784     & 3.2G         & 6673      & 9      \\ \hline
\end{tabular}
}
\end{table}

\subsubsection{Disaggregation with 
 neighbors}
\begin{figure}[htb]
	\label{fig_effect_of_distant_neighbors}
	\centering
	\includegraphics[scale=0.4]{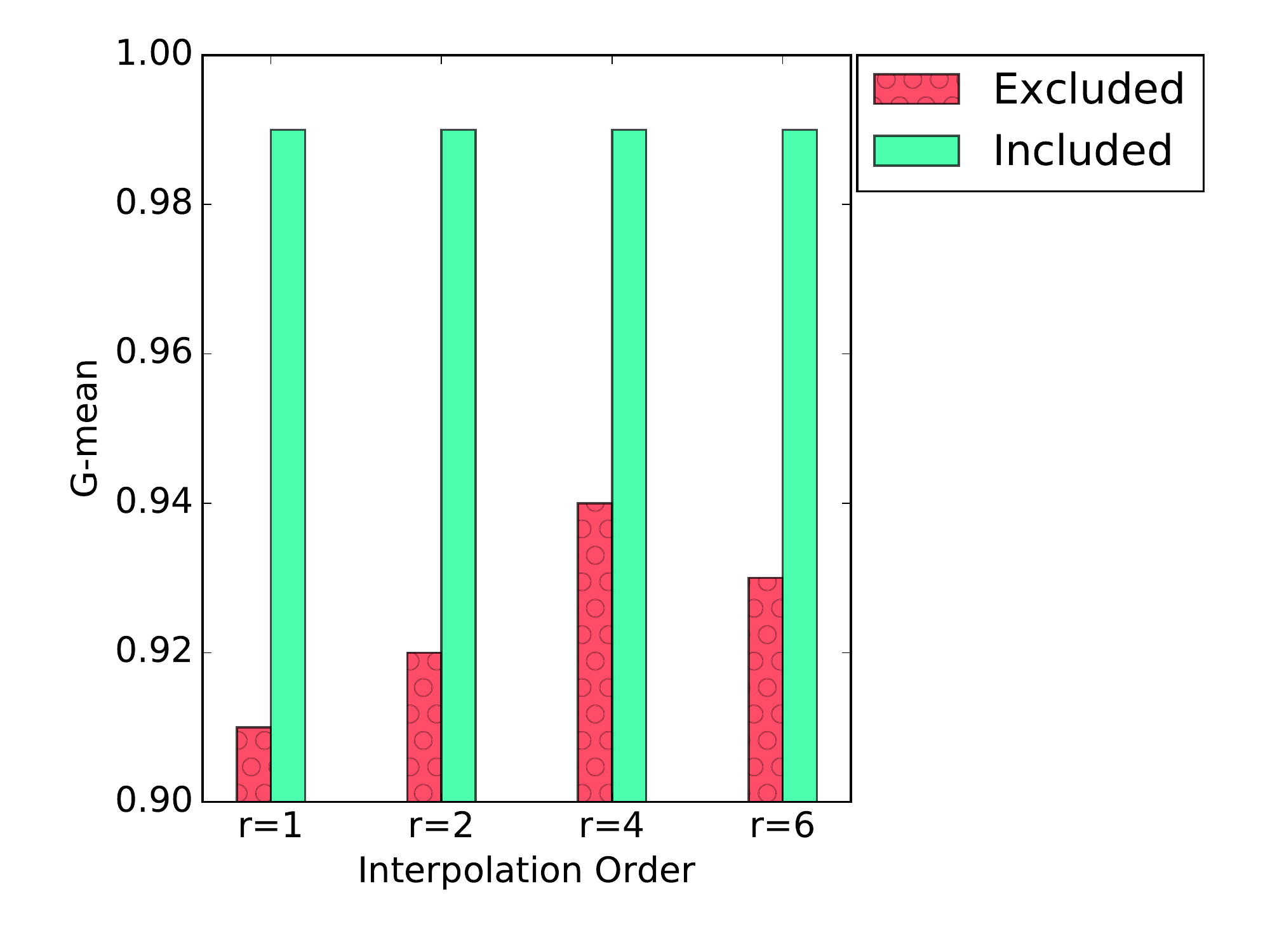}
	\caption{Effect of considering distance-1  disaggregation during the refinement phase on the G-mean for the Letter data set}
\end{figure}

When the computational resources allow and the k-NN graph is not extremely dense, one may add neighboring nodes to the corresponding disaggregated support vector nodes. While this adds flexibility to train the models (with more added data points), in most cases, it is an unnecessary step that increases the running time. The Forest, Clean, and Letter are the three data sets which demonstrate an improvement on classification quality by adding the distance-1 neighbors. 
The results for including the distant neighbors for the Letter data set experimenting with multiple coarse neighbor size reveal the largest improvement for $r=1$ (see Figure \ref{fig_effect_of_distant_neighbors}).

\subsubsection{Using partitioning in the refinement}
When the training set becomes too big during the refinement (at any level), a partitioning is used to accelerate the performance. In Table  \ref{tbl:forest_partitioning}, we compare the classification quality (G-mean), the size of training data, and the computational time.
In columns ``Partitioned'' (``Full''), we show these three factors when (no) partitioning is applied. 
When no partitioning is used, we train the model with the whole training data at each level. 
The partitioning starts when the size of training data is 5000 points. 
Typically, at the very coarse levels the size of training data is small, so in the experiment demonstrated in Table  \ref{tbl:forest_partitioning}, we show the numbers beginning level 5, the last level at which the partitioning was not applied.
The results in this and many other similar experiments show significant improvement in computational time when the training data is partitioned with very minor loss in  G-mean. The best level in the hierarchy is considered as the final level which is selected based on G-mean. Therefore, with no partitioning we obtain G-mean 0.79 and with partitioning it is 0.77 which are not significantly different results.

\begin{table}[H]
	\centering
	\caption{The G-mean, training set size, and computational time are reported for levels 1-5 of Forest data set for Class 5. The partitioning is started with 5000 points.}
	\label{tbl:forest_partitioning}
	\scalebox{.8}{
	\begin{tabular}{ccc|cc|cc}
    \hline
			  & \multicolumn{2}{c|}{G-mean } & \multicolumn{2}{c|}{Size of training set} & \multicolumn{2}{c}{Computational time}\\ \cline{2-7}
		Level & Full & Partitioned & Full & Partitioned  & Full   & Partitioned  \\
		\hline
		5 & 0.69 & 0.69 & 4387   & 4387  & 373   & 373  \\	
		4 & 0.68 & 0.72 & 18307  & 18673 & 1624  & 1262 \\
		3 & 0.79 & 0.77 & 47588  & 43977 & 6607  & 1528 \\
		2 & 0.79 & 0.72 & 95511  & 33763 & 17609 & 917  \\
		1 & 0.72 & 0.74 & 138018 & 24782 & 27033 & 576  \\
		\hline
	\end{tabular}
	}
\end{table}

\subsubsection{Comparison with fast Ensemble SVM}

A typical way to estimate the correctness of a multilevel solver is to compare its performance to those that use the local refinement techniques only. 
The EnsembleSVM \cite{claesen2014ensemblesvm} is a free software package containing efficient routines to perform ensemble learning with SVM models. The implementation exhibits very fast performance avoiding duplicate storage and evaluation of support vectors which are shared between constituent models. In fact, it is similar to our refinement and can potentially replace it in the multilevel framework. The comparison of our method with EnsembleSVM is presented in Table 
 \ref{ref_ensemblesvm_result_with_mlsvm}. While the running time is incomparable because of the obvious reasons (the complexity of EnsembleSVM is comparable to that of our last refinement only), the quality of our solver is significantly higher.

\begin{table}[H]
    \centering
    \caption{Ensemble SVM on benchmark data sets}
    \label{ref_ensemblesvm_result_with_mlsvm}
    \scalebox{.8}{
    \begin{tabular}{lcccc|cccc}
        \hline
        &
        \multicolumn{4}{c|}{Ensemble SVM} &
        \multicolumn{4}{c}{mlsvm-AMG} \\ \cline{2-9}    
                Dataset    & ACC   & SN    & SP    & G-mean             & ACC  & SN   & SP   & G-mean       \\ \hline
        Advertisement  & 0.52  & 0.41  & 0.95  & 0.57               & 0.95 & 0.85 & 0.96 & \bf{0.91}    \\
        Buzz           & 0.65  & 0.36  & 0.99  & 0.59               & 0.94 & 0.95 & 0.94 & \bf{0.95}    \\
        Clean (Musk)   & 0.85  & 0.00  & 0.85  & 0.00               & 0.99 & 0.99 & 0.99 & \bf{0.99}    \\
        Cod-RNA        & 0.90  & 0.82  & 0.94  & 0.88               & 0.93 & 0.97 & 0.91 & \bf{0.94}    \\
        Forest         & 0.98  & 0.32  & 0.99  & 0.57               & 0.77 & 0.96 & 0.80 & \bf{0.88}    \\
        Letter         & 0.97  & 0.75  & 0.98  & 0.86               & 0.98 & 0.99 & 0.98 & \bf{0.99}    \\
        Nursery        & 0.68  & 1.00  & 0.68  & 0.82               & 1.00 & 1.00 & 1.00 & \bf{1.00}    \\
        Ringnorm       & 0.68  & 0.61  & 1.00  & 0.78               & 0.98 & 0.98 & 0.98 & \bf{0.98}    \\
        Twonorm        & 0.75  & 0.89  & 0.76  & 0.81               & 0.98 & 0.98 & 0.97 & \bf{0.98}    \\
        \hline
    \end{tabular}
    }
\end{table}

\section{Conclusions}

 
In this paper we introduced novel multilevel frameworks for nonlinear support vector machines.
%
and discussed the details of several techniques for engineering multilevel frameworks that lead to a good trade-off between quality and running time. 
%
We ran a variety of experiments to compare several state-of-the-art SVM libraries and our frameworks on the classification quality and computation performance. 
The computation time of the proposed multilevel frameworks exhibits a significant improvement compared to the state-of-the-art SVM libraries with comparable or improved classification quality. For large data sets with more than 100,000 and up to millions of data points, we observed an improvement of computational time within an order of magnitude in comparison to DC-SVM and more two orders of magnitude in comparison to LibSVM. The improvement for larger datasets is even more significant.
The code for mlsvm-AMG is available at https://github.com/esadr/mlsvm.

There exist several attractive directions for the future research. One of them is to study in-depth why generating models at the coarse scales eliminates the effects of over- and under-fitting, a phenomena that we observed in many data sets. Another research avenue is to develop an uncoarsening scheme which chooses an appropriate kernel type at the coarse levels (where the training set size is relatively small) and continues with the best choice to fine levels. Indeed, if we successfully fit the parameters of kernel at the coarse levels, why not to try to choose the kernel type as well.
\pagebreak
\section*{Appendix A: Summary of parameters}

In Table \ref{tab:params2}, we mention recommended ranges of parameters for multilevel (W)SVM frameworks that we tested in our experiments.
\begin{table}[H]
    \centering
    \caption{Recommended parameter values.}
    \label{tab:params2}
    \scalebox{.8}{
    \begin{tabular}{|l|p{2cm}|p{7cm}|}
    \hline
    Parameter & Reference & Description\\
    \hline
        $r$ &  Sec. \ref{ref_amg_real_point} & Recommended range $[1,..,4]$. Almost all results were produced with $r=1$ except Cod-RNA ($r=2$) and SUSY ($r=4$).\\
        \hline
    $\theta$    & Sec. \ref{alg_amg_agg_point_coarsening} & Recommended range $[0.001,..,0.05]$. Almost all results were produced with $\theta=0.05$ except Letter ($\theta=0.005$) and Musk ($\theta=0.001$) that produced slightly better results with less aggressive filtering.\\
    \hline
    $d$ & Eq. (\ref{eq:multmodpred}) & Euclidean distance was used in all experiments.\\
         \hline
    $Q_t$ & Alg. \ref{alg:iisref} & Our simple single processor hardware allowed to start partitioning at 5000 data points. However, $Q_t$ in a range $[3000,..,5000]$ produced similar results.\\ 
    \hline
    $\eta$ & Alg. \ref{alg:coarsening} & In all experiments $\eta=2$.\\
   \hline
       $K$ & Alg. \ref{alg:iisref} & To preserve fast partitioning and training by parts, we used $K=\lfloor|\JJ_{(i)}|/1000\rceil$ for all levels $i$. No difference when changing this value was observed.\\     
   \hline
   $M^+$ and $M^-$ & Alg. \ref{alg:dr} & In all experiments $M^+ = M^- = 300$.\\
   \hline
   $|\JJ_{(\rho)}|$ & Alg. \ref{alg:dr} & The size of the coarsest level was always $|\JJ_{(\rho)}|$ = 500 to maintain fast performance of model selection at the coarsest level.\\
   \hline
   $Q$ & $\textsf{coarsen}$-IIS and $\textsf{coarsen}$-AMG (Alg. \ref{alg:coarsening}) & In all experiments $Q=0.5$. No significant difference was observed for $Q\in [0.4,...,0.6]$, see Fig. \ref{fig:scal_param_Q}.\\
   \hline
   $C$ and $\gamma$ & NUD in Alg. \ref{alg:iisref} & The NUD model selection algorithm starts parameter search in range of $2^{-10} < C <  2^{10}$ and $2^{-10} < \gamma <  2^{10}$ for the RBF kernel using the standard 9-13 scheme described in \cite{huang2007model}. \\
   \hline
    \end{tabular}
    }
\end{table}

\section*{Appendix B: Standard deviation for \mlsvm-AMG}
\begin{table}[H]
\centering
\caption{Standard deviations of the performance measures for \mlsvm-AMG}
\label{tbl:std}
  \scalebox{.8}{
    \begin{tabular}{lcccc}                                            \hline
    Dataset		         & ACC      & SN	   & SP       & G-mean   \\\hline
    Advertisement        & 0.01     & 0.04     & 0.01     & 0.02     \\ 
    Buzz 	             & 0.00       & 0.01     & 0.01     & 0.00       \\ 
    Clean (Musk)         & 0.00       & 0.00       & 0.00       & 0.00       \\ 
    Cod-RNA	             & 0.00       & 0.00       & 0.00       & 0.00       \\ 
    Forest   	         & 0.01     & 0.01     & 0.01     & 0.00       \\ 
    Letter   	         & 0.00       & 0.01     & 0.00       & 0.00       \\ 
    Nursery 	         & 0.00       & 0.00       & 0.00       & 0.00       \\ 
    Ringnorm 	         & 0.00       & 0.00       & 0.00       & 0.00       \\ 
    Twonorm 	         & 0.00       & 0.00       & 0.01     & 0.00       \\ 
    SUSY                 & 0.01     & 0.04     & 0.04     & 0.01     \\
    MNIST8M (Class 0)    & 0.01     & 0.00       & 0.01     & 0.01     \\ 
    MNIST8M (Class 1)    & 0.00       & 0.00       & 0.00       & 0.00       \\ 
    MNIST8M (Class 2)    & 0.02     & 0.02     & 0.02     & 0.02     \\ 
    MNIST8M (Class 3)    & 0.02     & 0.02     & 0.02     & 0.00       \\
    MNIST8M (Class 4)    & 0.02     & 0.01     & 0.02     & 0.01     \\
    MNIST8M (Class 5)    & 0.03     & 0.03     & 0.03     & 0.02     \\
    MNIST8M (Class 7)    & 0.02     & 0.02     & 0.02     & 0.02     \\
    MNIST8M (Class 6)    & 0.02     & 0.02     & 0.02     & 0.02     \\
    MNIST8M (Class 8)    & 0.03     & 0.03     & 0.04     & 0.01     \\
    MNIST8M (Class 9)    & 0.01     & 0.02     & 0.02     & 0.00       \\
    HIGGS                & 0.00       & 0.02     & 0.02     & 0.00       \\\hline
\end{tabular}
}
\end{table}

\section*{Acknowledgements}
We would like to thank three anonymous reviewers whose valuable comments helped to improve this paper significantly. This material is based upon work supported by the National Science Foundation under Grants No. 1638321 and 1522751.

\bibliographystyle{spmpsci}      


\end{document}